\documentclass[10pt,twocolumn,letterpaper]{article}

\usepackage[pagenumbers]{cvpr} 

\iftrue     
\newcommand{\dimpp}[1]{\textcolor{blue}{[DP: #1]}}
\newcommand{\vicky}[1]{\textcolor{magenta}{ \textbf{#1}}}

\newcommand{\thanos}[1]{\textcolor{orange}{[AT: #1]}}

\else
\newcommand{\dimpp}[1]{\textcolor{blue}{\noindent}}
\newcommand{\vicky}[1]{\textcolor{magenta}{\noindent}}
\newcommand{\thanos}[1]{\textcolor{orange}{\noindent}}
\fi

\newcommand{\mypar}[1]{\vspace{0mm}\noindent\textbf{#1}}

\newcommand{\myparb}[1]{\vspace{0mm}\noindent\textbf{#1}}

\definecolor{ourgreen}{RGB}{46, 204, 113}
\definecolor{ourred}{RGB}{231, 76, 60}
\definecolor{ourwhite}{RGB}{236, 240, 241}

\usepackage{subcaption}

\usepackage{booktabs}
\usepackage{xspace} 
\usepackage{booktabs}
\usepackage{color, colortbl}
\usepackage{multirow}
\usepackage{comment}
\usepackage{arydshln}
\usepackage{xspace} 
\usepackage[accsupp]{axessibility}

\usepackage{graphicx,overpic}
\usepackage{amsmath, amssymb}
\usepackage[super]{nth}
\usepackage{nicematrix}
\usepackage{tabularx,stackengine,collcell}
\let\endminwd\relax
\newcolumntype{L}[1]{>{\collectcell\xminwd l{#1}}l<{\endminwd\endcollectcell}}
\newcolumntype{C}[1]{>{\collectcell\xminwd c{#1}}c<{\endminwd\endcollectcell}}
\newcolumntype{R}[1]{>{\collectcell\xminwd r{#1}}r<{\endminwd\endcollectcell}}
\def\minwd#1#2#3\endminwd{\stackengine{0pt}{#3}{\rule{#2}{0pt}}{O}{#1}{F}{F}{L}}
\newcommand\xminwd[1]{\minwd#1}

\usepackage{pifont}
\newcommand{\cmark}{\ding{51}}%
\newcommand{\xmark}{\ding{55}}%
\newcommand{\mj}{$\mathcal{J}$}
\newcommand{\mf}{$\mathcal{F}$}
\newcommand{\mjf}{$\mathcal{J}\&\mathcal{F}$}

\usepackage{tikz}

\definecolor{cvprblue}{rgb}{0.21,0.49,0.74}
\usepackage[pagebackref,breaklinks,colorlinks,allcolors=cvprblue]{hyperref}

\def\paperID{6} 
\def\confName{CVPR}
\def\confYear{2025}

\title{Studying Image Diffusion Features for Zero-Shot Video Object Segmentation}

\author{Thanos Delatolas$^{1,2}$ \quad 
Vicky Kalogeiton$^3$ \quad 
Dim~P.~Papadopoulos$^{1,2}$ \quad \\
$^{1}$\,Technical University of Denmark \quad
$^{2}$\,Pioneer Center for AI\\
$^{3}$\,LIX, Ecole Polytechnique, CNRS,  Institut Polytechnique de Paris 
\\
{\tt\small atde@dtu.dk, vicky.kalogeiton@polytechnique.edu, dimp@dtu.dk}
\\\url{https://diff-zsvos.compute.dtu.dk/}
}

\begin{document}
\maketitle
\begin{abstract}
This paper investigates the use of large-scale diffusion models for Zero-Shot Video Object Segmentation (ZS-VOS) without fine-tuning on video data or training on any image segmentation data. While diffusion models have demonstrated strong visual representations across various tasks, their direct application to ZS-VOS remains underexplored. Our goal is to find the optimal feature extraction process for ZS-VOS  by identifying the most suitable time step and layer from which to extract features. We further analyze the affinity of these features and observe a strong correlation with point correspondences. Through extensive experiments on DAVIS-17 and MOSE, we find that diffusion models trained on ImageNet outperform those trained on larger, more diverse datasets for ZS-VOS. Additionally, we highlight the importance of point correspondences in achieving high segmentation accuracy, and we yield state-of-the-art results in ZS-VOS. Finally, our approach performs on par with models trained on expensive image segmentation datasets.
\end{abstract}  
\section{Introduction}
\label{sec:intro}
\begin{figure}[t]
    \centering
    \includegraphics[width=\linewidth]{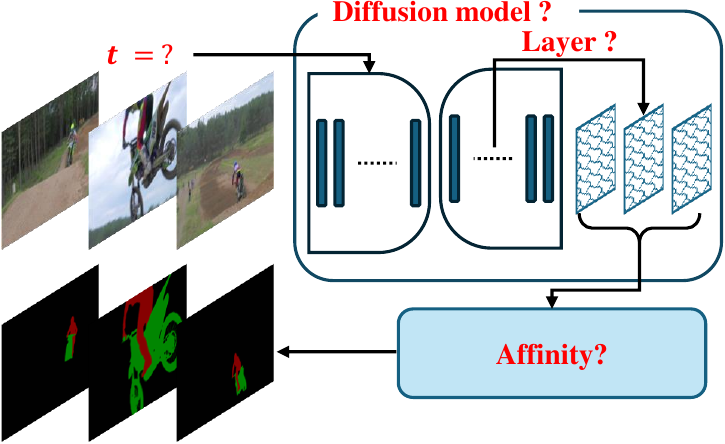}
    \caption{
    We leverage pre-trained diffusion models for Zero-Shot Video Object Segmentation by addressing key challenges: selecting the appropriate \emph{diffusion model}, determining the optimal \emph{time step}, identifying the best feature extraction \emph{layer}, and designing an effective \emph{affinity} matrix calculation strategy to match the features.
    }
    \label{fig:teaser}
\end{figure}

Large-scale diffusion models trained on vast datasets have demonstrated exceptional capabilities~\cite{Rombach_2022_CVPR, stable_diffusion_3} in text-to-image and text-to-video generation~\cite{esser2023structure}. These models learn rich representations, making them attractive for adaptation to discriminative tasks such as semantic segmentation~\cite{vpd,xu2023odise,namekata2024emerdiff,zhang2024three}, point correspondences~\cite{dift, zhang2023tale,Zhang_2024_CVPR,stracke2024cleandift}, depth estimation~\cite{ke2023repurposing,saxena2023surprising}, and video segmentation~\cite{alimohammadi2024smite,chen2023generalist,zhu2024exploring}. However, most prior work relies on finetuning~\cite{vpd,ke2023repurposing,zhang2024three}, limiting their zero-shot applicability. Adapting these representations for downstream tasks without additional training remains an open challenge~\cite{namekata2024emerdiff}. 

Semi-supervised Video Object Segmentation (VOS) is the task of segmenting objects in videos given their first-frame segmentation mask.
State-of-the-art VOS methods~\cite{xmem,bekuzarov2023,cutie} are trained on large-scale video datasets~\cite{mose,youtube_vos}, yet their performance drops significantly on more challenging benchmarks~\cite{mose, vost}. This highlights the limitations of supervised training on fixed datasets. 
Scaling up VOS training is impractical due to the cost of annotating segmentation masks for every video frame~\cite{coco,delatolas2024learning,delatolas2023eva}. Additionally, existing VOS models rely on ResNet~\cite{he2016deep} features pre-trained on ImageNet~\cite{russakovsky2015imagenet}, which may be suboptimal due to their supervised learning paradigm. In contrast, large-scale diffusion models~\cite{Rombach_2022_CVPR} are trained with self-supervised objectives, capturing richer and more diverse representations~\cite{stracke2024cleandift}, offering better feature representations for VOS.

In this paper, we explore how to leverage features from pre-trained image diffusion models for Zero-Shot VOS (ZS-VOS)~\cite{jabri2020space,meng2024segic,fragkiadaki2015learning} without any finetuning on video data or training on any image segmentation data. This eliminates the need for costly video and image annotations while reducing computational overhead. However, directly using diffusion features for VOS presents key challenges, including identifying the most informative representations and ensuring reliable temporal correspondences. We systematically address these challenges, demonstrating that diffusion models are powerful feature extractors for ZS-VOS without any task-specific finetuning. 

To adapt diffusion models for ZS-VOS, we address two key challenges. First, we identify the most suitable features by selecting the optimal time step and layer (Fig.~\ref{fig:teaser}). Since the time step in the diffusion process strongly influences internal representations~\cite{stracke2024cleandift}, and different layers encode varying levels of semantic information~\cite{ldznet}, choosing the optimal combination is crucial. Secondly, we delve deeper into the affinity of these features~\cite{dift}, which predicts segmentation masks through frame-to-frame correspondences. We enhance feature matching by filtering incorrect correspondences and introducing a prompt-learning strategy~\cite{peng2024harnessing} that leverages the text prompt of Stable Diffusion~\cite{Rombach_2022_CVPR}.

Extracting useful knowledge from large diffusion models is non-trivial~\cite{zhang2024three}. Through extensive experiments on DAVIS-17~\cite{davis_17} and MOSE~\cite{mose}, we identify several findings:
(a) All versions of stable diffusion yield the best segmentation accuracy when extracting features from the same layer.
(b) The Ablated Diffusion Model (ADM)~\cite{adm}, trained on ImageNet~\cite{russakovsky2015imagenet}, significantly outperforms all versions of Stable Diffusion~\cite{Rombach_2022_CVPR}, despite being smaller in size. 
(c) Incorrect point correspondences significantly impact performance, highlighting the need for precise feature matching in the VOS task. 
(d) We achieve state-of-the-art ZS-VOS performance without \emph{any} training on video data or pretraining using image segmentation annotations. Our approach yields performance comparable to models trained on large image segmentation datasets (such as SA-1B~\cite{kirillov2023segment}). 

\section{Related Work}
\label{sec:relwork}
\myparb{Semi-supervised Video Object Segmentation (VOS)} segments objects given their first-frame
segmentation mask. Early methods~\cite{on_1,on_2,on_3,on_4,on_5,on_6} overfit networks at test-time but suffer from high computational cost. Propagation-based methods~\cite{prop_1,prop_2,prop_3,prop_4,prop_5,prop_6,prop_7} perform frame-to-frame propagation, resulting in faster runtime. However, they cannot capture long-term context and struggle with occlusions and appearance changes. 
Memory-based methods~\cite{stm,xie2021efficient,stcn,xmem,bekuzarov2023,cutie,zhou2024rmem} use a memory bank of previous frames with their corresponding predictions and perform pixel-level matching between the memory frames and the current frame. 
Transformer-based methods~\cite{aot,deaot,Wang_2023_CVPR,Wu_2023_ICCV_VOS,goyal2024tamvt,li2024learning} enable object-level reasoning using variants of attention to reduce the space/time complexity.
Unlike these VOS-specific approaches trained on multiple VOS datasets, our analysis focuses on the zero-shot version the semi-supervised VOS task.

\myparb{Zero-shot VOS (ZS-VOS)} evaluates the generalization ability of models to the semi-supervised VOS task without finetuning on video data or training on any image segmentation data~\cite{uziel2023vit,dift,dino}. Apart from this paradigm, many approaches have tested zero-shot capabilities by relying on image segmentation datasets~\cite{wang2023seggpt,zou2023segment,wang2023images,meng2024segic} or relying only on unlabeled video data~\cite{li2023unified,invos,jabri2020space}.

\begin{figure*}
    \centering
    \includegraphics[width=\linewidth]{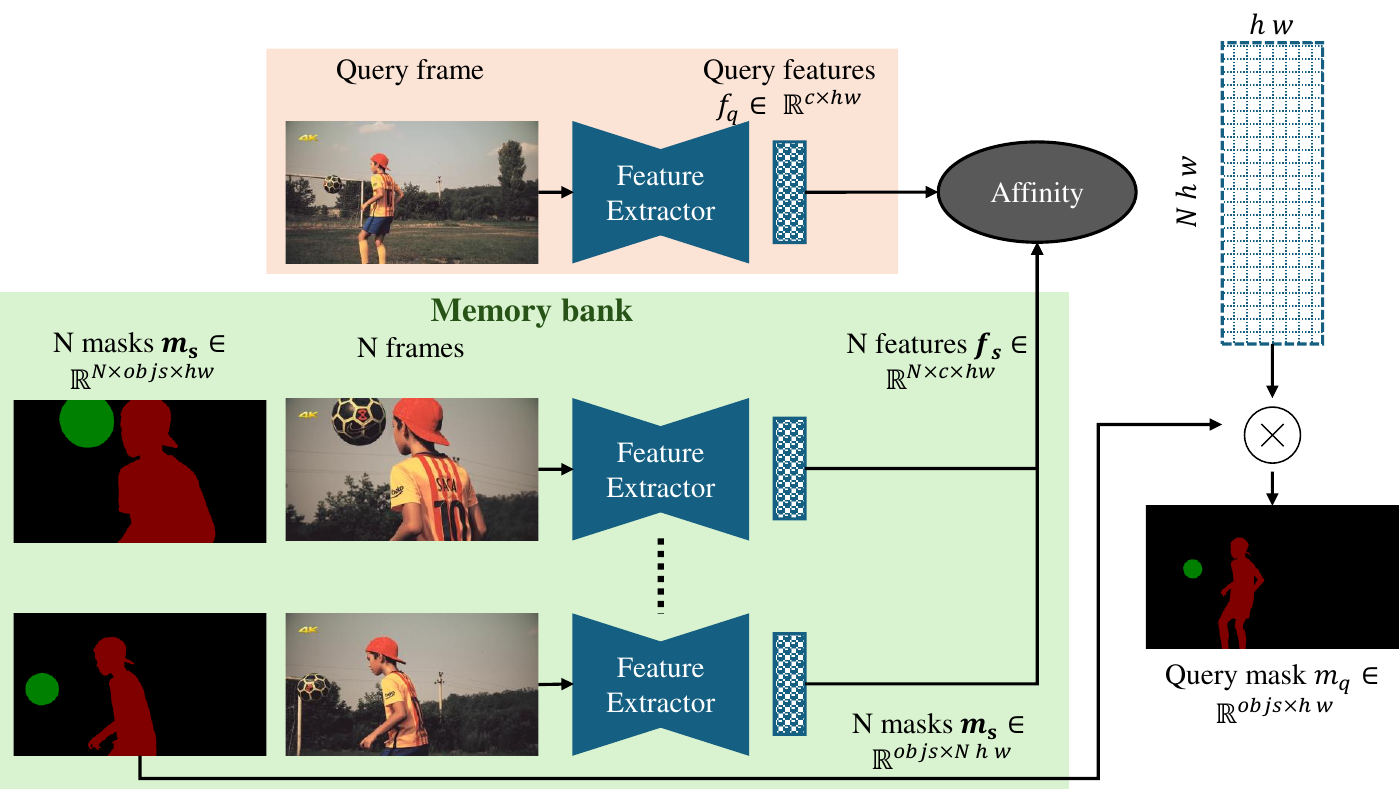}
    \caption{\textbf{Sequentially segmenting a video with powerful feature extractors~\cite{dino,Rombach_2022_CVPR} and past predictions.} Given a memory of $N$ past frames and their corresponding predicted segmentation masks, we segment the query frame by first calculating the affinity matrix $\mathcal{A}$ between the query and memory frames, and then multiplying $\mathcal{A}$ with the past predicted segmentation masks.}
    \label{fig:fig2}
\end{figure*}

\myparb{Diffusion models} are generative models~\cite{ddpm,song2020score} trained to gradually denoise images. The Ablated Diffusion Model~\cite{adm} outperformed GANs in image synthesis on ImageNet~\cite{russakovsky2015imagenet}. To reduce the required computational resources while achieving state-of-the-art results, Stable Diffusion~\cite{Rombach_2022_CVPR} was introduced, where it is trained to gradually denoise the latents of a VAE~\cite{Esser_2021_CVPR}. Stable Diffusion builds upon classifier-free guidance~\cite{ho2022classifier} and generates images given a text caption. It is trained on billions of text-image pairs from LAION~\cite{schuhmann2022laion}. More recently, DiT~\cite{dit} introduced a transformer-based denoiser, replacing the de facto U-Net~\cite{ronneberger2015u}. Building on DiT, Stable Diffusion~3~\cite{stable_diffusion_3} proposes incorporating flow-matching~\cite{lipman2022flow} during training.

\myparb{Diffusion features for discriminative tasks.} Many approaches have leveraged diffusion features for discriminative tasks. They can be categorized into four groups: (1) training conditional diffusion models to generate annotations~\cite{amit2021segdiff,Chen_2023_ICCV,chen2023generalist}, (2) generating synthetic image-annotation pairs to train a decoder~\cite{Wu_2023_ICCV, wu2023datasetdm}, (3) fine-tuning large-scale diffusion models for a downstream task~\cite{vpd, xu2023odise, namekata2024emerdiff, zhang2024three, alimohammadi2024smite,kondapaneni2024tadp}, and (4) 
leveraging diffusion features directly or optimizing minimal parameters at test time~\cite{dift, gong2023prompting, zhang2024diff}. In the video domain, Pix2Seq-D~\cite{chen2023generalist} treats panoptic segmentation as a generative task but does not leverage large-scale diffusion models~\cite{stable_diffusion_3,Rombach_2022_CVPR}. 
VD-IT~\cite{zhu2024exploring} trains a matching framework on video data, consisting of CLIP~\cite{clip} and DETR~\cite{detr}, while SMITE~\cite{alimohammadi2024smite} finetunes the cross-attention layers of Stable Diffusion~\cite{Rombach_2022_CVPR}. Diff-Tracker~\cite{zhang2024diff} avoids large-scale video training by introducing a test-time prompt learning strategy for video tracking. Inspired by this, we introduce a similar prompt learning for ZS-VOS, achieving segmentation accuracy comparable to ground-truth text.

\section{Method}
\label{sec:method}
In this work, we examine powerful diffusion models as feature extractors~\cite{Rombach_2022_CVPR, adm, stable_diffusion_3} for the task of ZS-VOS, without finetuning on video or training on image segmentation data. More formally, given a video $ V = \{I_1, I_2, \ldots, I_K\}$ consisting of $ K$ frames, and the ground-truth mask of the first frame, $ m_1 \in \mathbb{R}^{\text{objs} \times h w}$, where $\text{objs}$ is the number of objects in the video and $h$ and $w$ are the height and width of the frames, we sequentially segment the remaining frames.

To do this, we maintain a memory bank of the $N$ most recent predicted segmentation masks $\mathbf{m}_s$, and their corresponding frames (Fig.~\ref{fig:fig2}). The memory is initialized with the ground-truth mask of $m_1$ and $I_1$. To segment a new query frame, we first extract features (Sec.~\ref{method:feats}) for both the memory and query frames. Then, we calculate the affinity matrix $\mathcal{A}$ between the memory and query features, which represents how much each memory pixel corresponds to each query pixel. Finally, to predict the segmentation mask of the query frame, $m_q$, we multiply the past predictions, $\mathbf{m}_s$, by the affinity matrix $\mathcal{A}$ (Sec.~\ref{method:feature_mathching_mask_prediction}).
Given that $\mathcal{A}$ estimates how much each pixel in memory corresponds to each query pixel, we propose improving these correspondences to enhance the segmentation quality of $m_q$ (Sec.~\ref{method:correspondences}).

\begin{figure*}
    \centering
    \includegraphics[width=\linewidth]{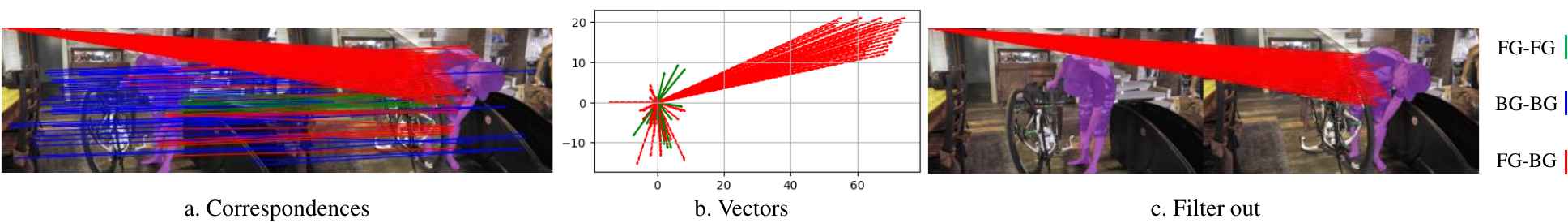}
    \caption{\textbf{Correspondences.} (a) We show the FG-FG, BG-BG, and FG-BG correspondences. (b) We show the vectors of correspondences in the cartesian space. (c) We filter out the correspondences with our MAG-Filter.}
    \label{fig:corrs_method}
\end{figure*}

\begin{figure}
    \centering
    \includegraphics[width=\linewidth]{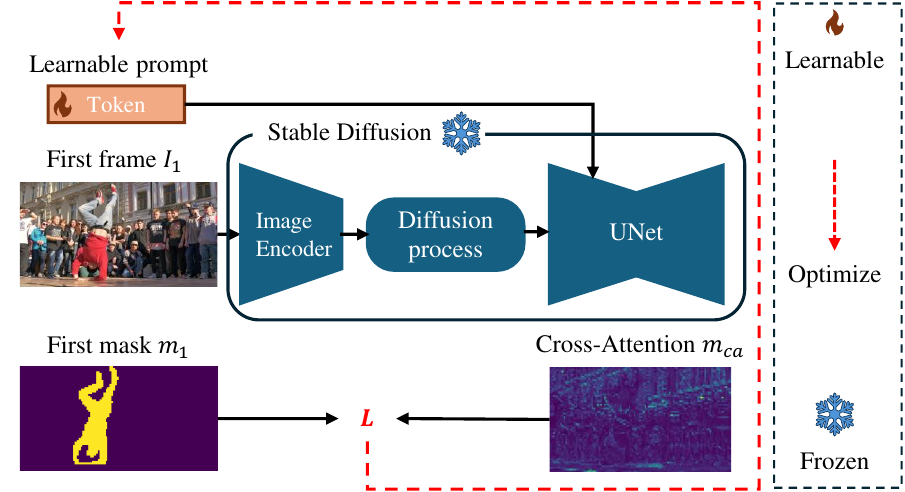}
    \caption{\textbf{Prompt Learning strategy in ZS-VOS.} Given the first frame of the video, $I_1$, and its corresponding segmentation mask, $m_1$, we optimize a text token so that its cross-attention map, $m_{ca} $, approximates $m_1 $.}
    \label{fig:prompt_learning}
\end{figure}

\subsection{Preliminaries}
\label{method:theory}
We briefly review diffusion models. Diffusion models~\cite{ddpm,adm} are trained to gradually denoise a noisy image, $x_t$. The noisy image is created from the clean image $x_0$ using:

\begin{equation}
    x_t = \sqrt{a_t} x_0 + \sqrt{1 - a_t} \epsilon
    \label{eq:forward}
\end{equation}
\noindent
where $a_t$ depends on the time step $t$ and blends the noise with the image, and $\epsilon \sim \mathcal{N}(0,1)$ is the Gaussian noise. The time step $t$ determines the noise level, with $t=0$ corresponding to the clean image and $t=T$ corresponding to pure noise. A neural network, $ g_\theta $, typically a UNet~\cite{ronneberger2015u}, takes as input the noisy image $ x_t $, the time step $ t $, and optionally a conditioning input $ c $, and is trained to predict the noise $ \epsilon $. The condition $c$ can be a text description of the image, a segmentation mask, or any other input relevant to the clean image $x_0$. Once $ g_\theta $ is trained, it can generate images by gradually refining an initial pure noise input $ x_T $.

\subsection{Feature Extraction}
\label{method:feats}
Given a video frame, we extract a feature map using diffusion models~\cite{adm, Rombach_2022_CVPR, stable_diffusion_3}. Since diffusion models are trained to denoise images, we first generate a noisy version of the frame, $\hat{I}$, at a specific time step $t$ using Eq.~\eqref{eq:forward}. Then, following prior work~\cite{dift, vpd, xu2023odise, stracke2024cleandift}, we feed $\hat{I}$ into the U-Net~\cite{ronneberger2015u} of the diffusion model and extract a feature map from its intermediate decoder layers. If the diffusion model is conditioned on text, we prompt it with an empty string. 
\subsection{Feature matching and mask prediction}
\label{method:feature_mathching_mask_prediction} Given a memory bank with the $N$ most recent predicted segmentation masks, $\mathbf{m_s} \in \mathbb{R}^{\text{objs} \times N \times h w}$, the features of the corresponding frames $\mathbf{f}_\mathbf{s} \in \mathbb{R}^{N \times c \times h w}$, and the features of the query frame $f_q \in \mathbb{R}^{c \times h w}$, our goal is to predict the segmentation mask of the query frame  $\mathbf{m_q} \in \mathbb{R}^{\text{objs} \times h w}$. To achieve this, we first compute the affinity matrix $\mathcal{A}$ between the memory and query features and then predict $m_q$ by multiplying $\mathcal{A}$ with $\mathbf{m_s}$.

\myparb{Affinity matrix.} The affinity matrix $\mathcal{A}$ indicates the correlation between each memory pixel and each pixel of the query. We compute $ \mathcal{A} \in \mathbb{R}^{N h w \times h w} $, using the following similarity functions between the memory features, $\mathbf{f}_\mathbf{s}$, and the query features, $ f_q $:

\noindent
\textit{Cosine (COS)}: $ \mathcal{A} = \mathbf{f}_\mathbf{s} ^T \cdot f_q$,
where each feature vector in $ f_q $ and $ \mathbf{f}_\mathbf{s} $ is L2-normalized along the channel dimension.

\noindent
\textit{L1}: $\mathcal{A} = -\sum_{c=1}^{C} \left| \mathbf{f}_\mathbf{s}^{(c)} - f_q^{(c)} \right|$,
%
where the sum is taken over all channels $ c $. The negative sign ensures that higher values indicate more similarity.

\noindent
\textit{L2}: $\mathcal{A} = -\sqrt{\sum_{c=1}^{C} \left( \mathbf{f}_\mathbf{s}^{(c)} - f_q^{(c)} \right)^2}$,
%
where again, the sum is taken over all channels $c$. The negative sign ensures that higher values indicate more similarity.

\begin{figure*}[htbp]
    \centering
    \begin{subfigure}[b]{0.32\textwidth}
        \centering
        \includegraphics[width=\textwidth]
        {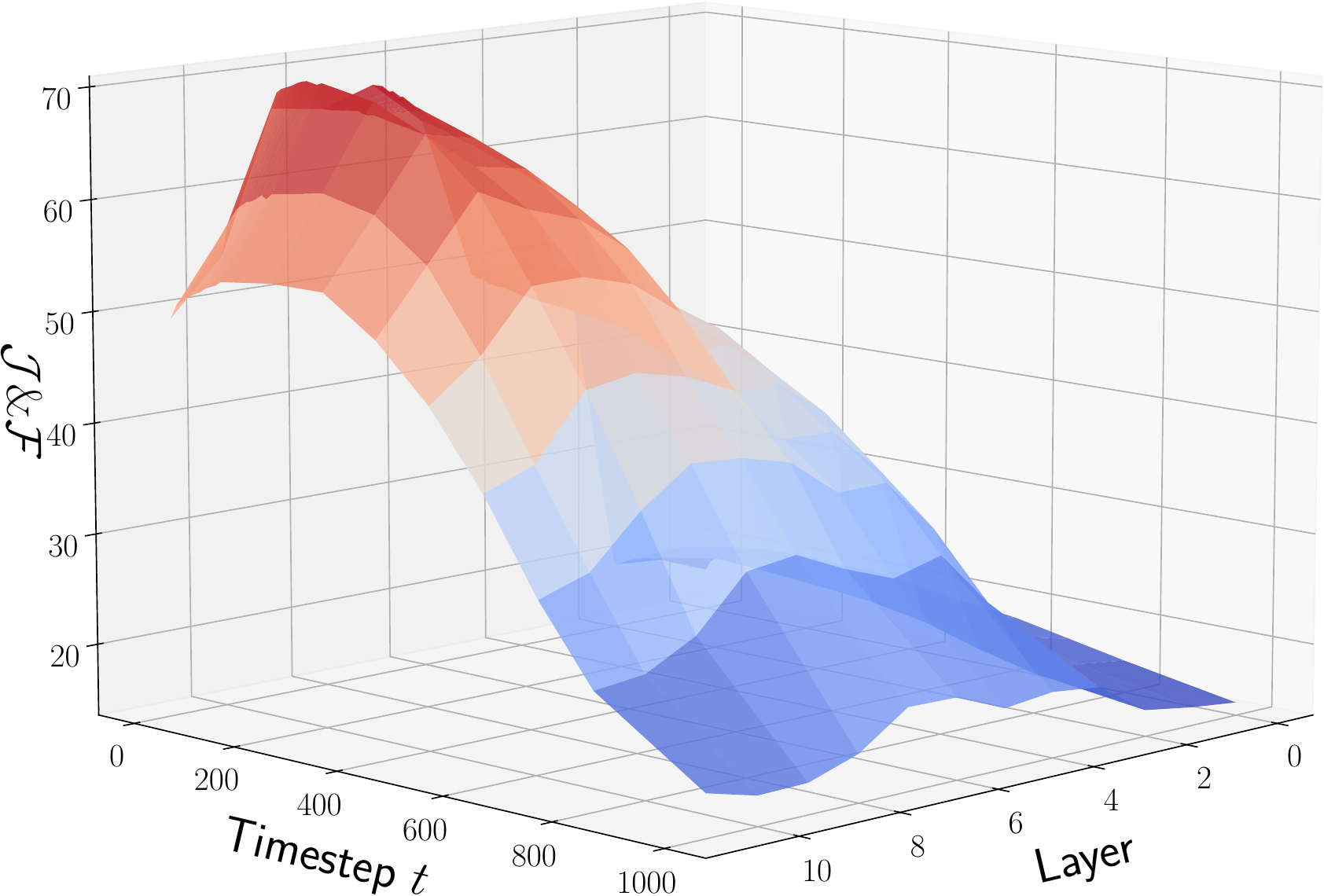}
        \caption{Stable Diffusion 2.1}
        \label{fig:sd_21}
    \end{subfigure}
    \begin{subfigure}[b]{0.32\textwidth}
        \centering
        \includegraphics[width=\textwidth]{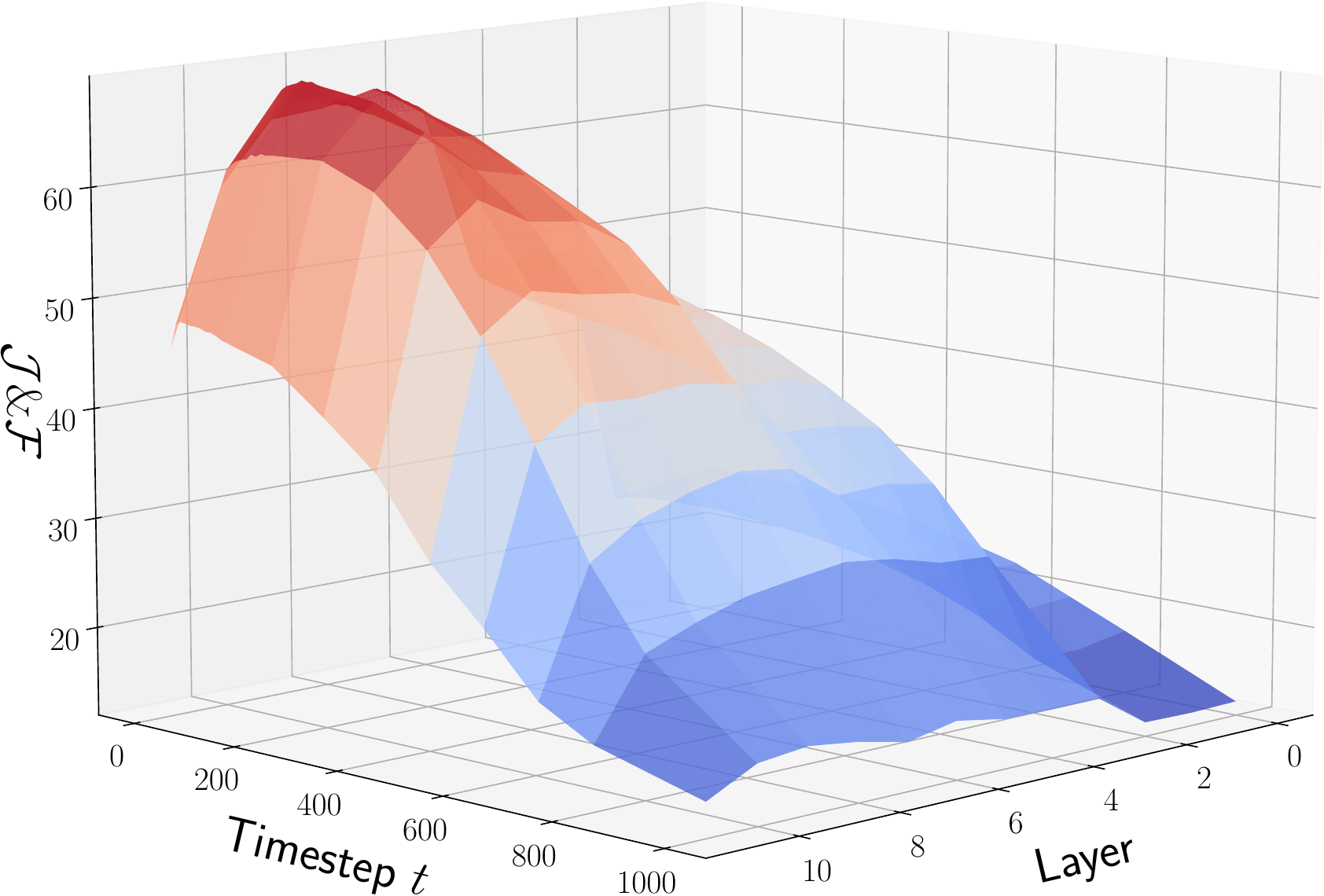}  
        \caption{Stable Diffusion 1.5}
        \label{fig:sd_15}
    \end{subfigure}
    \begin{subfigure}[b]{0.32\textwidth}
        \centering
        \includegraphics[width=\textwidth]{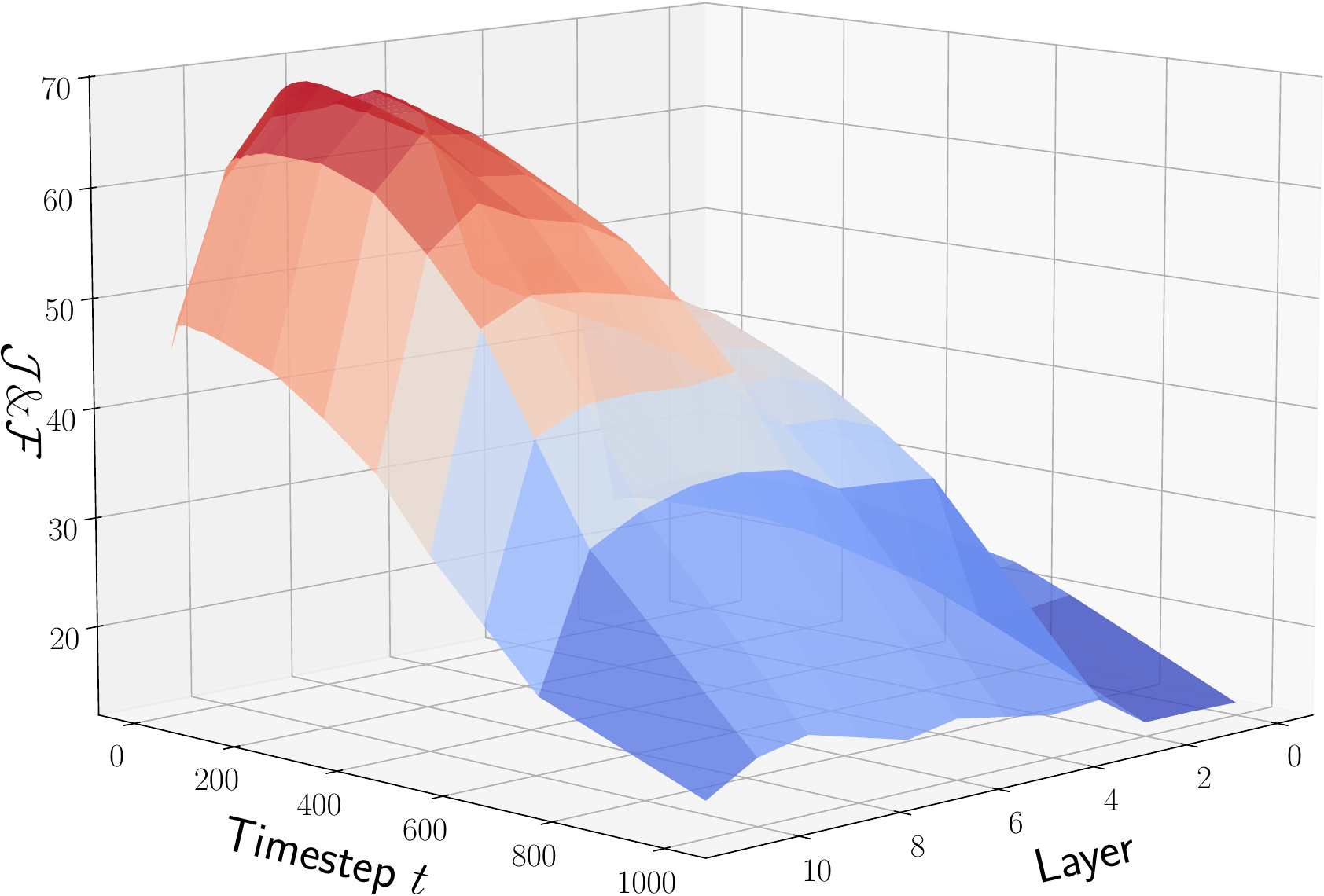} 
        \caption{Stable Diffusion 1.4}
        \label{fig:sd_14}
    \end{subfigure}
    \begin{subfigure}[b]{0.32\textwidth}
        \centering
        \includegraphics[width=\textwidth]{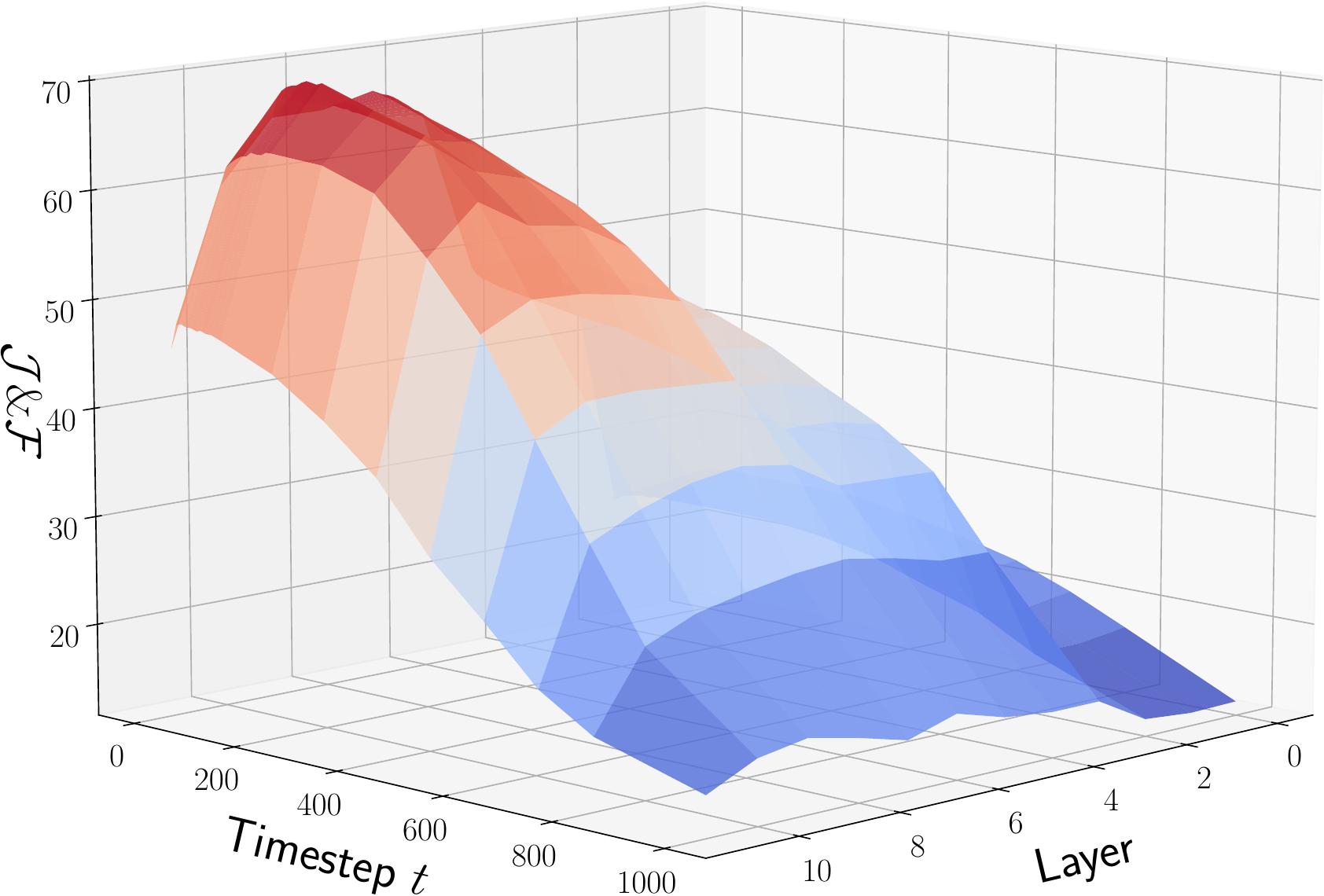} 
        \caption{Stable Diffusion 1.3}
        \label{fig:sd_13}
    \end{subfigure}
    \begin{subfigure}[b]{0.32\textwidth}
        \centering
        \includegraphics[width=\textwidth]{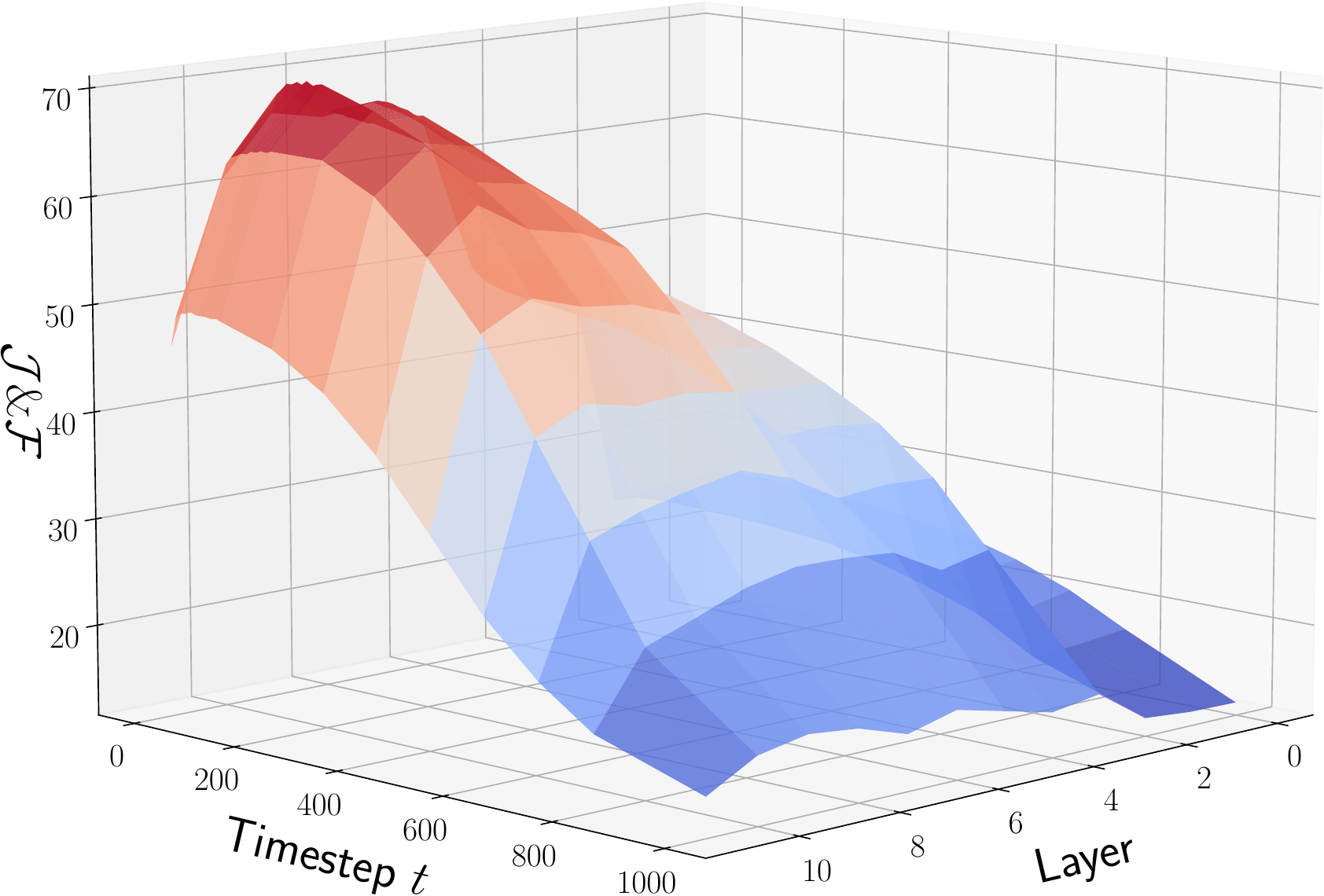}  
        \caption{Stable Diffusion 1.2}
        \label{fig:sd_12}
    \end{subfigure}
    \begin{subfigure}[b]{0.32\textwidth}
        \centering
        \includegraphics[width=\textwidth]{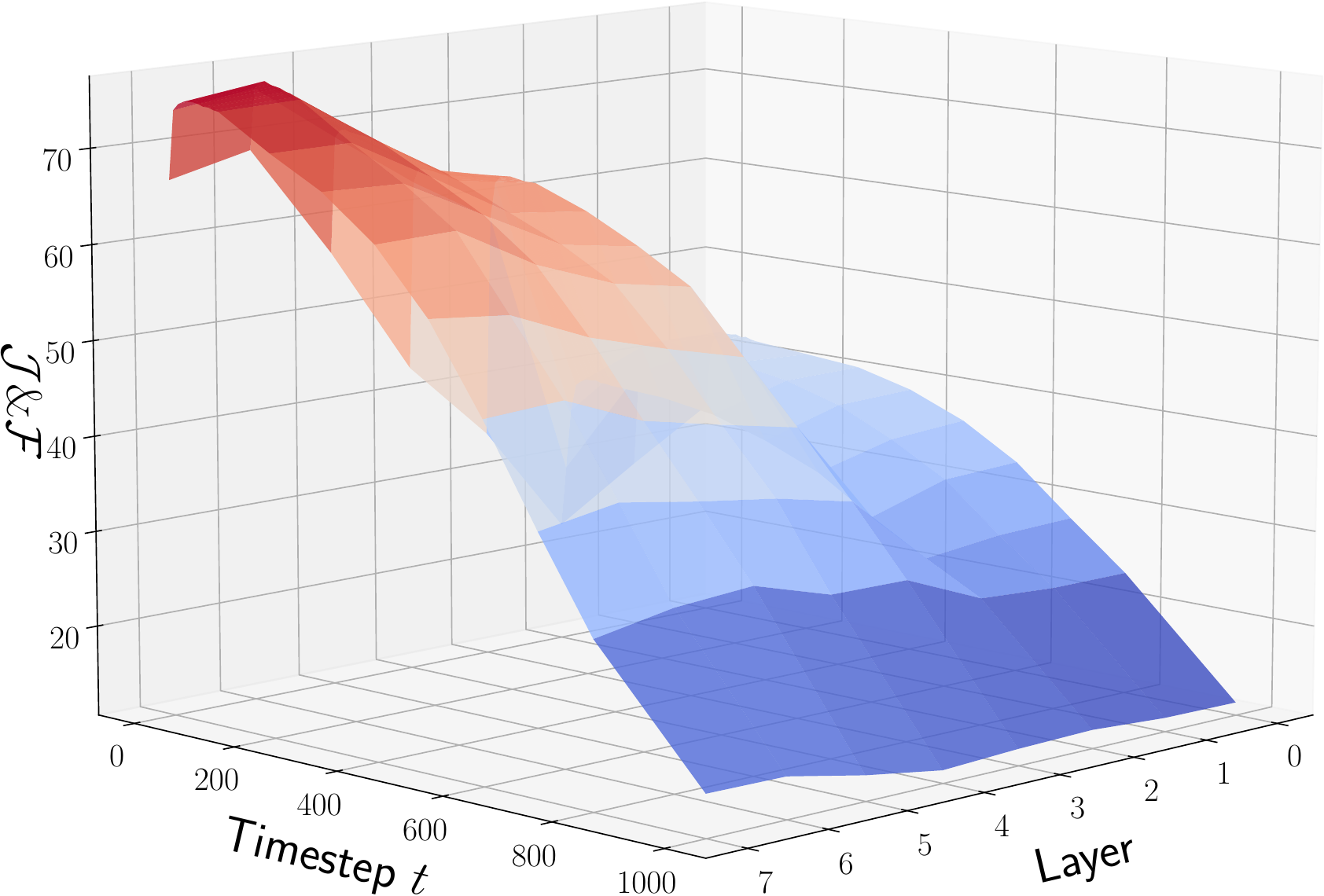}
        \caption{Ablated Diffusion Model (ADM)}
        \label{fig:sd_adm}
    \end{subfigure}
    \caption{\textbf{Ablation on layer and time step.} We show the \mjf\xspace accuracy on the DAVIS-17 val set~\cite{davis_17} for Stable Diffusion (v 1.2 to 1.5 and 2.1), as well as the Ablated Diffusion Model (ADM)~\cite{adm}, as a function of the diffusion time step and the decoder layer of the U-Net.}
    \label{fig:ablation_timestep_layer}
\end{figure*}

\subsection{Improving Correspondences}
\label{method:correspondences}
Given, two frames $I_1$ and $I_2$, their affinity matrix $\mathcal{A} \in \mathbb{R}^{hw \times hw}$ indicates how much each pixel of $I_1$ corresponds to $I_2$. We compute the point correspondences of $ I_1 $ to $ I_2 $ by taking the maximum of the affinity matrix $ \mathcal{A} $ over the second dimension of $ I_2 $. Specifically, for each pixel in $ I_1 $, we find the pixel in $ I_2 $ that has the highest affinity:

\[
\text{correspondence}(i) = \arg \max_j \mathcal{A}(i, j)
\]
\noindent
where $ i $ indexes the pixels of $ I_1 $ and $ j $ indexes the pixels of $ I_2 $, and the result gives us the index of the most corresponding pixel in $ I_2 $ for each pixel in $ I_1 $.

\begin{figure}[t]
    \centering
    \includegraphics[width=\linewidth]{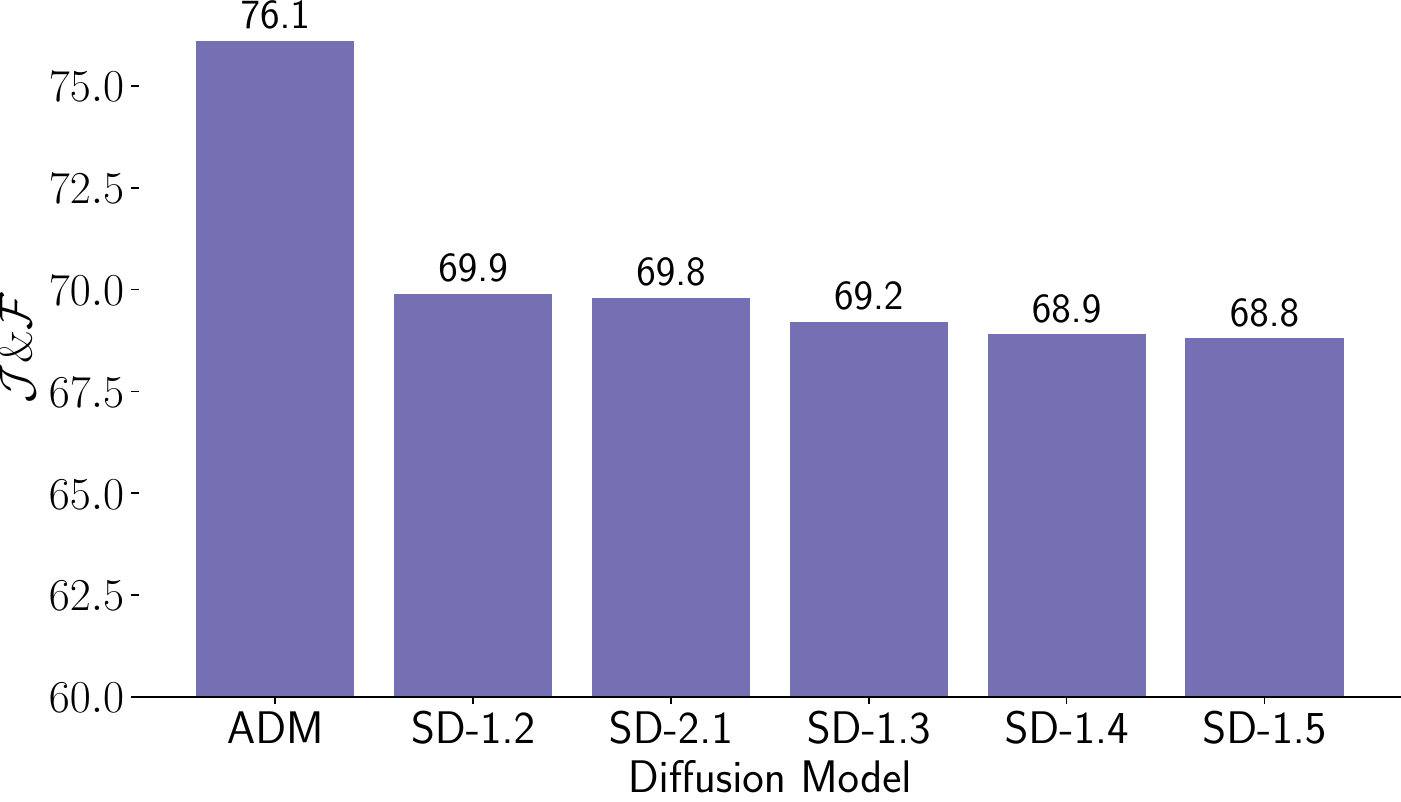}
    \caption{\textbf{Highest \mjf\xspace across all layers and timepsteps} for each diffusion model on the DAVIS-17~\cite{davis_17} val set.}
    \label{fig:best_jand_diffmodels}
\end{figure}

\myparb{Categories of correspondences.} In Fig.~\ref{fig:corrs_method}(a), we illustrate the correspondences between the first and the twentieth frames of a video sequence from DAVIS-17~\cite{davis_17}. We categorize them into three types: foreground-foreground (FG-FG), foreground-background (FG-BG), and background-background (BG-BG). Given two ground-truth masks,  $m_1 $ and $ m_2$, we define three correspondence categories as follows:
\noindent
\textit{Foreground-Foreground (FG-FG)}: A correspondence is considered FG-FG if it belongs to the foreground in both frames, as indicated by the following mask:
\begin{equation}
    \text{fg\_fg\_mask} = m_1 \land m_2
\end{equation}

\noindent
\textit{Background-Background (BG-BG)}: A correspondence is categorized as BG-BG if it belongs to the background in both frames:
\begin{equation}
    \text{bg\_bg\_mask} = (\neg m_1) \land (\neg m_2)
\end{equation}

\noindent
\textit{Foreground-Background (FG-BG)} A correspondence falls into the FG-BG category if it transitions between foreground and background across frames. We compute this by identifying pixels that are foreground in one frame but background in the other:
\begin{equation}
    \begin{split}
        \text{fg\_bg1} &= m_1 \land (\neg m_2) \\
        \text{fg\_bg2} &= (\neg m_1) \land m_2 \\
        \text{fg\_bg\_mask} &= \text{fg\_bg1} \lor \text{fg\_bg2}    
    \end{split}
\end{equation}

\noindent
FG-BG correspondences represent incorrect or mismatched correspondences and are considered wrong because they indicate a pixel that transitions from foreground in one frame to background in the other, or vice versa.

\myparb{FG-BG percentage.} We define the FG-BG percentage as the proportion of the top $ k $ correspondences in the affinity matrix that belong to the $ \text{fg\_bg\_mask} $. A lower percentage is preferable as it indicates fewer mistakenly identified foreground-background correspondences.

\myparb{Magnitude filter (MAG-Filter).}
In Fig.~\ref{fig:corrs_method}(b), we plot the correspondence vectors in the Cartesian system between the first and fortieth frames with the highest affinity values from the bike-packing video in DAVIS-17\cite{davis_17}. For each pixel$(i,j)$ that corresponds to$(\hat{i}, \hat{j})$, we calculate the vector $\vec{v} = (\hat{i}-i, \hat{j}-j)$, which indicates the direction and magnitude for each correspondence. We observe that some FG-BG vectors have higher magnitudes than FG-FG and BG-BG vectors. Thus, we filter out correspondences with a magnitude higher than $r$. In Fig.~\ref{fig:corrs_method}(c), we show the filtered correspondences, and we observe that some FG-BG are filtered out, but no FG-FG correspondences are.

\myparb{Prompt Learning.} Given that text-to-image diffusion models~\cite{stable_diffusion_3, Rombach_2022_CVPR} build mappings (correspondences) between text and images, we leverage this to create improved image-to-image correspondences~\cite{peng2024harnessing, khani2023slime, gong2023prompting, zhang2024diff}. Fig.~\ref{fig:prompt_learning} illustrates our prompt learning strategy. Since the ground-truth mask $m_1$ of the first frame $I_1$ is provided at test time, we optimize a token so that its cross-attention map $m_{ca}$ approximates $m_1$. In the case of a video with multiple objects, we learn one token per object. For the loss function $L$, we experiment with MSE, BCE, MSE with the diffusion loss, and BCE with the diffusion loss (DM)~\cite{peng2024harnessing}.

\section{Experiments}
\label{sec:exps}
We present our analysis on zero-shot semi-supervised video object segmentation (ZS-VOS) without finetuning on video or training on image segmentation data. We analyze and justify all design choices on DAVIS-17~\cite{davis_17} and validate our findings on additional datasets.
We first identify the most suitable features and similarity functions across multiple models (Sec.~\ref{exps:diff_feats_analysis}). Then, we evaluate our MAG-filter and prompt learning strategy to improve the correspondences (Sec.~\ref{exps:corrs_analysis}). Finally, we compare against state-of-the-art VOS methods (Sec.~\ref{exps:zero_shot_vos}) and validate our findings on additional datasets (Sec.~\ref{exps:gen_ability}).

\subsection{Experimental settings}
\myparb{Datasets.} \textbf{DAVIS-17} provides high-quality annotated masks and is split into 60 training, 30 validation, and 30 test videos. We use the validation set of DAVIS-17 for our analysis and also report performance on the test set. \textbf{MOSE} contains 2,149 videos, split into 1,507 training, 311 validation, and 331 testing videos. MOSE is one of the most challenging datasets, as it contains many objects, heavy occlusions, and the appearance-reappearance of objects. We report performance on the val set using the evaluation server~\footnote{https://codalab.lisn.upsaclay.fr/competitions/10703}.

\myparb{Implementation details.} We conduct our experiments using the Ablated Stable Diffusion (ADM)~\cite{adm}, Stable Diffusion (SD) versions~\cite{Rombach_2022_CVPR} 1.2 to 1.5 and 2.1, as well as DINO~\cite{dino}. The total time step  $T$ for all diffusion models is 1000. Following prior work~\cite{dift, alimohammadi2024smite, vpd}, we extract features from the decoder of the UNet. In particular, ADM's UNet has 18 decoder layers, but we extract features from the first eight due to computational constraints. SD's UNet consists of 4 decoder layers, each having 3 ResNet~\cite{he2016deep} blocks. To analyze the decoder's features, we extract features from all ResNet blocks and refer to each output as a different layer. Unless stated otherwise, SD is prompted with an empty string. We use the base version of DINO~\cite{dino} trained on ImageNet~\cite{russakovsky2015imagenet}, and we extract a features from the last layer of the ViT~\cite{vit}. We remove the $[CLS]$ token and reshape the output features into a feature map. Following DIFT~\cite{dift}, we use the original 480p version for all datasets in all models. Finally, we use $r=25\sqrt{2}$ in the MAG-filter.
We evaluate segmentation quality using the Jaccard index \mj, contour accuracy \mf, and their average \mjf~\cite{perazzi2016benchmark}.

\begin{figure}[t]
    \centering
    \includegraphics[width=\linewidth]{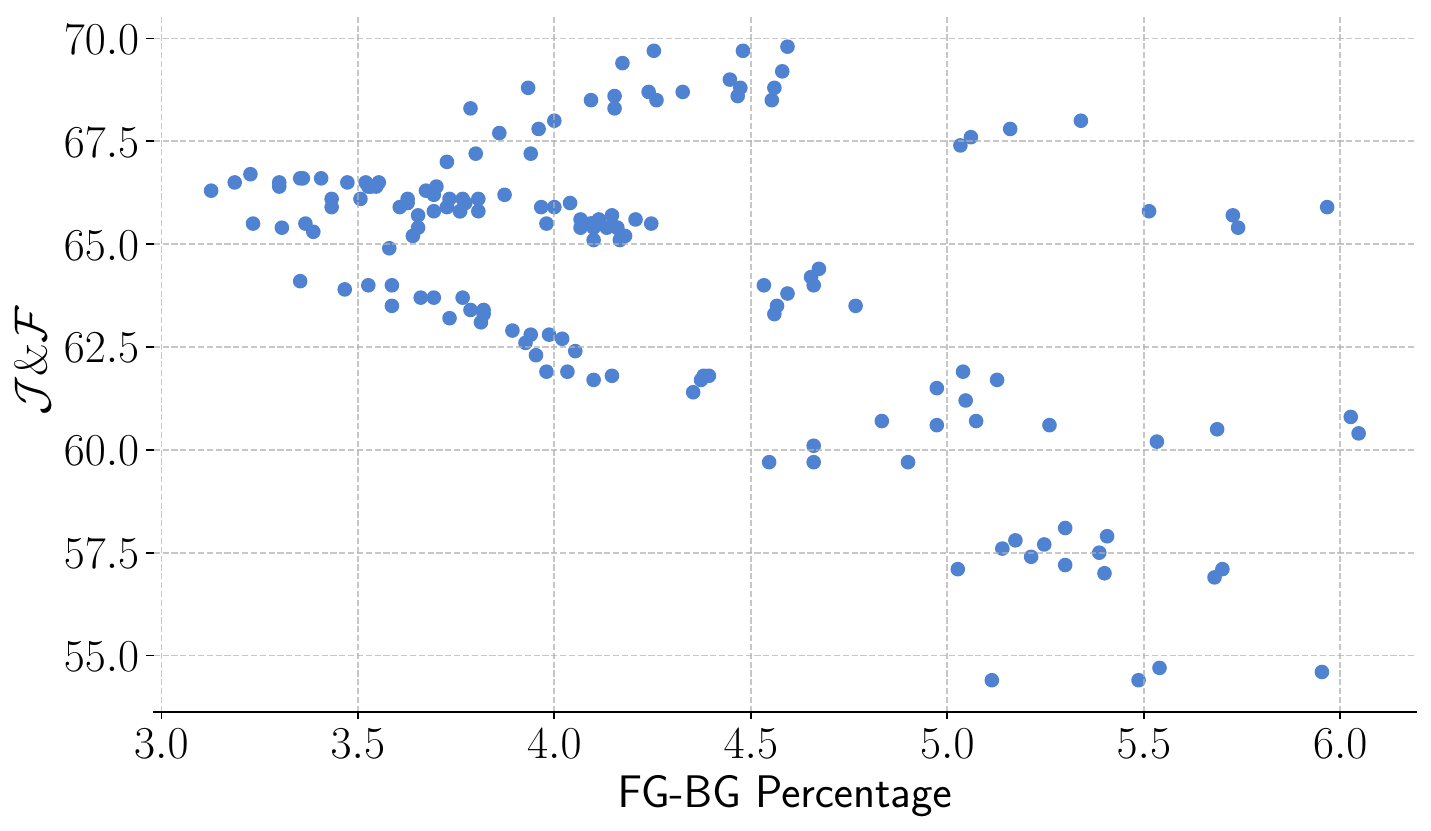}
    \caption{\textbf{FG-BG percentage vs \mjf.} We show the FG-BG percentage in comparison to the \mjf\xspace on the DAVIS-17 val set across the Stable Diffusion~\cite{Rombach_2022_CVPR} versions 1.2 to 1.5, as well as 2.1.}
    \label{fig:fg_bg_jandf}
\end{figure}

\begin{table}[t]
    \centering
\small
\begin{NiceTabular}
{lc@{\hspace{10pt}}C{2.2em}@{}C{2.2em}@{}C{2.2em}}[colortbl-like]
\toprule
& & \multicolumn{3}{c}{\small DAVIS-17 val} \\
\cmidrule(lr){3-5} 
Model  & Affinity &  \mjf & \mj & \mf \\
\midrule
\multirow{3}{*}{DINO~\cite{dino}} & \textit{L1} &  8.0 &7.6 &8.4 \\
& \textit{COS} & 71.4 & 68.0 & 74.7 \\
& \textit{L2} & \textbf{71.6} &\textbf{68.2} & \textbf{75.0}\\
\hdashline
\multirow{3}{*}{SD 2.1~\cite{Rombach_2022_CVPR}} & \textit{L1} & 65.6 & 62.0& 69.3\\
& \textit{COS} & \textbf{69.8} & \textbf{67.1} & 72.6\\
& \textit{L2} & \textbf{69.8} &66.9 & \textbf{72.7}\\
\hdashline
\multirow{3}{*}{ADM~\cite{adm}} & \textit{L1} & 55.9  &53.4 &58.4 \\
& \textit{COS} & 76.1& \textbf{73.2}& 79.1\\
& \textit{L2} &\textbf{76.2}  &\textbf{73.2} & \textbf{79.2}\\
\bottomrule
\end{NiceTabular}
    \caption{\textbf{Affinities.} We compare different similarity metrics for the affinity matrix for DINO~\cite{dino}, SD 2.1~\cite{Rombach_2022_CVPR}, and ADM~\cite{adm}.}
    \label{tab:affinity_ablation}
\end{table}


\subsection{Diffusion features analysis}
\label{exps:diff_feats_analysis}
We begin our analysis by identifying the best features for ZS-VOS using DIFT~\cite{dift} on the DAVIS-17~\cite{davis_17} val set.

\begin{table}[t]
    \centering
\small
\begin{NiceTabular}
{lc@{\hspace{10pt}}C{4.2em}@{}C{4.2em}@{}C{4.2em}}[colortbl-like]
\toprule
& & \multicolumn{3}{c}{\small DAVIS-17 val} \\
\cmidrule(lr){3-5} 
Model  & MAG-Filter &  \mjf & \mj & \mf \\
\midrule
\multirow{2}{*}{SD-1.2} & \color{ourred}\xmark & \multicolumn{1}{l}{69.9} & \multicolumn{1}{l}{67.3}  & \multicolumn{1}{l}{72.6}   \\
&\color{ourgreen}\cmark &  70.3\textsubscript{\color{ourgreen}$\blacktriangle $0.4}& 67.5\textsubscript{\color{ourgreen}$\blacktriangle $0.2} & 73.1\textsubscript{\color{ourgreen}$\blacktriangle $0.5}  \\ 
\hdashline
\multirow{2}{*}{SD-1.3} & \color{ourred}\xmark & \multicolumn{1}{l}{69.2} & \multicolumn{1}{l}{66.5}  & \multicolumn{1}{l}{72.0}   \\
&\color{ourgreen}\cmark &  69.5\textsubscript{\color{ourgreen}$\blacktriangle $0.3} &  66.6\textsubscript{\color{ourgreen}$\blacktriangle $0.1}   &  72.4\textsubscript{\color{ourgreen}$\blacktriangle $0.4}\\ 
\hdashline
\multirow{2}{*}{SD-1.4} & \color{ourred}\xmark & \multicolumn{1}{l}{68.9} & \multicolumn{1}{l}{66.0}  & \multicolumn{1}{l}{71.8}   \\
&\color{ourgreen}\cmark &  69.2\textsubscript{\color{ourgreen}$\blacktriangle $0.3} &  66.1\textsubscript{\color{ourgreen}$\blacktriangle $0.1}   &  72.1\textsubscript{\color{ourgreen}$\blacktriangle $0.5}    \\
\hdashline
\multirow{2}{*}{SD-1.5} & \color{ourred}\xmark & \multicolumn{1}{l}{68.8} & \multicolumn{1}{l}{66.0}  & \multicolumn{1}{l}{71.6}   \\
&\color{ourgreen}\cmark &   69.3\textsubscript{\color{ourgreen}$\blacktriangle $0.5} &  66.5\textsubscript{\color{ourgreen}$\blacktriangle $0.5} &   72.1\textsubscript{\color{ourgreen}$\blacktriangle $0.5}  \\
\hdashline
\multirow{2}{*}{SD-2.1} & \color{ourred}\xmark & \multicolumn{1}{l}{69.8} & \multicolumn{1}{l}{67.1}  & \multicolumn{1}{l}{72.6}  \\
& \color{ourgreen}\cmark &  70.2\textsubscript{\color{ourgreen}$\blacktriangle $0.4} &  67.2\textsubscript{\color{ourgreen}$\blacktriangle $0.1} &  73.1\textsubscript{\color{ourgreen}$\blacktriangle $0.5} \\
\hdashline
\multirow{2}{*}{ADM} & \color{ourred}\xmark & \multicolumn{1}{l}{76.1} &\multicolumn{1}{l}{73.2}  & \multicolumn{1}{l}{79.1}  \\
&\color{ourgreen}\cmark &   76.8\textsubscript{\color{ourgreen}$\blacktriangle $0.7} &  73.8 \textsubscript{\color{ourgreen}$\blacktriangle $0.6}   &   79.7\textsubscript{\color{ourgreen}$\blacktriangle $0.7}  \\ 
\hdashline
\multicolumn{2}{c}{Oracle} &  \textbf{83.3}\textsubscript{\color{ourgreen}$\blacktriangle $13.5} &  \textbf{77.8}\textsubscript{\color{ourgreen}$\blacktriangle $10.7} &  \textbf{88.9}\textsubscript{\color{ourgreen}$\blacktriangle $16.3} \\ 
\bottomrule
\end{NiceTabular}
    \caption{\textbf{Filter correspondences of the affinity matrix.} \textbf{Bold} denotes the best performing setting. We show in green the performance increase with respect to the default no filtering approach.}
    \label{tab:filter_correspondences}
\end{table}

\begin{table}[t]
    \centering
\small
\begin{NiceTabular}
{lc@{\hspace{10pt}}C{2.2em}@{}C{2.2em}@{}C{2.2em}}[colortbl-like]
\toprule
& & \multicolumn{3}{c}{\small DAVIS-17 val} \\
\cmidrule(lr){3-5} 
 Prompt  & Loss &  \mjf & \mj & \mf \\
\midrule
 empty text & -  & 69.8 &66.9 & 72.7\\
 GT-text & -  & \textbf{70.2}& \textbf{67.3} & \textbf{73.1} \\
 learnable & BCE  & \textbf{70.2} &67.2 &\textbf{73.1 }\\
 learnable & BCE+DM  & 69.9& 67.0&72.8\\
 learnable & MSE  & 70.1 & 67.2& \textbf{73.1}\\
 learnable & MSE+DM  & 70.1 &67.3 &73.0 \\
\bottomrule
\end{NiceTabular}
    \caption{\textbf{Prompt learning.} We compare SD 2.1~\cite{Rombach_2022_CVPR} prompted with an empty text, the ground-truth text, and our prompt learning strategy. \textbf{Bold} denotes the best performing setting.}
    \label{tab:prompt_learning}
\end{table}

\begin{figure*}
    \centering
    \includegraphics[width=\linewidth]{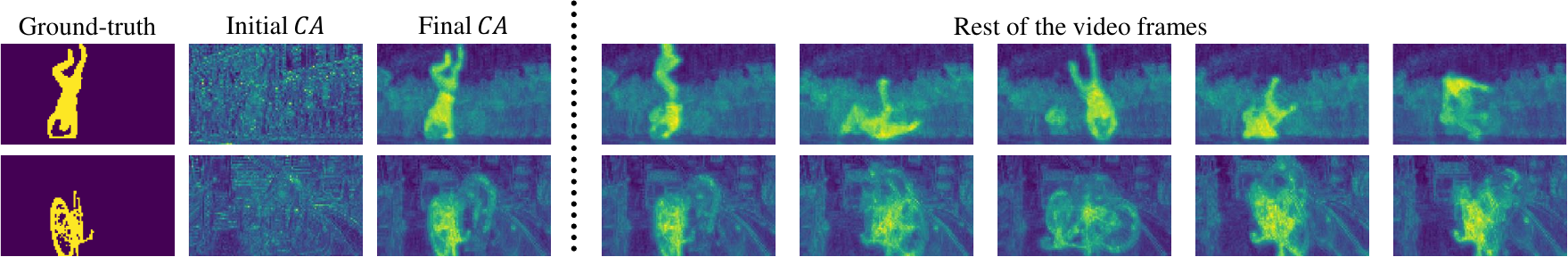}
    \caption{\textbf{Qualitative examples of Prompt Learning.} (Left) Cross-attention maps, $CA$, of SD-2.1~\cite{Rombach_2022_CVPR} before and after our prompt learning strategy. (Right) Cross-attention maps with the optimized token from the first frame.}
    \label{fig:prompt_learning_qualitative}
\end{figure*}

\begin{table*}
    \centering
\small
\begin{NiceTabular}
{lccccccc}[colortbl-like]
\toprule
 & \multicolumn{2}{c}{\small{Image-level}} & \multicolumn{2}{c}{\small{Video-level}} & & \multicolumn{1}{c}{\small{DAVIS-17 val}} \\
\cmidrule(lr){2-3}  \cmidrule(lr){4-5} \cmidrule(lr){7-7}
Model  & \#Images &\#Segmentations & \#Frames &\#Segmentations & Datasets & \mjf \\
\midrule
XMem~\cite{xmem} & 1.02M &  27K &150K&210K & I+S+D+Y &86.2\\
Cutie~\cite{cutie} & 1.02M &  27K &150K&210K & I+S+D+Y &88.8\\
SAM2~\cite{ravi2024sam} & 11M & 1.1B &4.2M&35.5M & SA+SAV &\textbf{90.7}\\
\hdashline
SegIC~\cite{meng2024segic}& 1.3M & 1.8M & \color{ourred}\xmark & \color{ourred}\xmark & I+C+A+L& 73.7\\
SegGPT~\cite{wang2023seggpt}& 147K & 1.62M & \color{ourred}\xmark & \color{ourred}\xmark & C+A+V&75.6\\
PerSAM-F~\cite{zhang2023personalize}& 11M & 1.1B & \color{ourred}\xmark & \color{ourred}\xmark & SA&76.1\\
Matcher~\cite{liu2023matcher}& 11M & 1.1B & \color{ourred}\xmark & \color{ourred}\xmark & SA&\textbf{79.5}\\
\hdashline
FGVG~\cite{li2023learning}&1M&\color{ourred}\xmark &116K&\color{ourred}\xmark&I+Y+FT&72.4\\
STT~\cite{stt}&1M& \color{ourred}\xmark &95K& \color{ourred}\xmark &I+Y &\textbf{74.1}\\
\hdashline
STC~\cite{jabri2020space}&\color{ourred}\xmark&\color{ourred}\xmark&20M&\color{ourred}\xmark&K&67.6\\
INO~\cite{invos}&\color{ourred}\xmark&\color{ourred}\xmark&20M&\color{ourred}\xmark&K&72.5\\
Mask-VOS~\cite{li2023unified}&\color{ourred}\xmark& \color{ourred}\xmark &95K& \color{ourred}\xmark &Y &\textbf{75.6}\\
\hdashline
MoCo~\cite{he2019moco}&1M& \color{ourred}\xmark &\color{ourred}\xmark& \color{ourred}\xmark &I &65.4\\
SHLS~\cite{Santos_2023_BMVC}&10K& \color{ourred}\xmark &\color{ourred}\xmark& \color{ourred}\xmark &M &68.5\\
DIFT-SD~\cite{dift}&5B& \color{ourred}\xmark &\color{ourred}\xmark& \color{ourred}\xmark & LN&70.0\\
DINO~\cite{dino}&1M& \color{ourred}\xmark &\color{ourred}\xmark& \color{ourred}\xmark & I &71.4\\
DIFT-ADM~\cite{dift}&1M& \color{ourred}\xmark &\color{ourred}\xmark& \color{ourred}\xmark & I & 75.7\\
Training-Free-VOS~\cite{uziel2023vit}&1M& \color{ourred}\xmark &\color{ourred}\xmark& \color{ourred}\xmark & I & 76.3\\
\cellcolor{ourwhite}SD-2.1+Prompt Learning&\cellcolor{ourwhite}5B& \cellcolor{ourwhite}\color{ourred}\xmark &\cellcolor{ourwhite}\color{ourred}\xmark& \cellcolor{ourwhite}\color{ourred}\xmark & \cellcolor{ourwhite}LN&\cellcolor{ourwhite}70.5\\
\cellcolor{ourwhite}ADM+MAGFilter&\cellcolor{ourwhite}1M& \cellcolor{ourwhite}\color{ourred}\xmark &\cellcolor{ourwhite}\color{ourred}\xmark& \cellcolor{ourwhite}\color{ourred}\xmark &\cellcolor{ourwhite} I&\cellcolor{ourwhite}\textbf{76.8}\\
\bottomrule
\end{NiceTabular}
    \caption{\textbf{Video Object Segmentation results.} We categorize state-of-the-art methods based on whether they are pre-trained on image-level or video-level data and/or fine-tuned on object segmentation annotations.
    Key for \textit{Datasets} column: I:ImageNet~\cite{russakovsky2015imagenet}, S=Static images that VOS models pretrain~\cite{wang2017learning,shi2015hierarchicalECSSD,zeng2019towardsHRSOD,cheng2020cascadepsp,li2020fss}, D=DAVIS-17~\cite{davis_17}, Y=YouTube~\cite{youtube_vos}, M=MSRA10K~\cite{msr30k}, C=COCO~\cite{coco}, A=ADE20k~\cite{ade20k_1,ade20k_2}, L=LVIS~\cite{gupta2019lvis}, V=VOC~\cite{voc}, SAV=SA-V~\cite{ravi2024sam}, SA=SA-1B~\cite{kirillov2023segment}, K=Kinetics~\cite{kay2017kinetics}, LN=LAION~\cite{schuhmann2022laion}, FT=FlyingThings~\cite{Mayer_2016_CVPR}.}
    \label{tab:zero_shot_vos_comp}
\end{table*}

\myparb{Layer and time step.} In Fig.~\ref{fig:ablation_timestep_layer}, we show \mjf\xspace as a function of layer and time step for SD (v 1.2 to 1.5 and 2.1) and ADM~\cite{adm}. We observe that earlier layers ($\leq 5$) and a high time step ($\geq 300$) yield low \mjf\xspace accuracy, peaking at 40\%. None of the models peak in performance at $t = 0$, which indicates that a small amount of noise is proper for feature extraction, as the UNet is trained for denoising. All SD versions~\cite{Rombach_2022_CVPR} peak in performance at layer 9, whereas in ADM~\cite{adm}, \mjf\xspace increases as the layer number increases.

\begin{table*}
    \centering
\small
\begin{NiceTabular}
{lcc@{\hspace{10pt}}C{2.2em}@{}C{2.2em}@{}C{2.2em}@{\hspace{5pt}}C{2.2em}@{}C{2.2em}@{}C{2.2em}@{\hspace{5pt}}C{2.2em}@{}C{2.2em}@{}C{2.2em}}[colortbl-like]
\toprule
&  &  & \multicolumn{3}{c}{\small MOSE} & \multicolumn{3}{c}{\small DAVIS-17 val} & \multicolumn{3}{c}{\small DAVIS-17 test}  \\
\cmidrule(lr){4-6}  \cmidrule(lr){7-9}  \cmidrule(lr){10-12} 
Model & Affinity & Prompt Learning & \mjf & \mj & \mf & \mjf & \mj & \mf & \mjf & \mj & \mf  \\
\midrule
\multirow{2}{*}{DINO~\cite{dino}} & \textit{COS} & - &33.7&\textbf{28.7}&38.8&  71.4 &68.0 & 74.7 & 63.3 & 57.7&68.9 \\
& \textit{L2}& - &\textbf{33.8}&\textbf{28.7}&\textbf{39.0} &  \textbf{71.6} &\textbf{68.2} & \textbf{75.0} &\textbf{63.5}& \textbf{57.8}&\textbf{69.1} \\
\hdashline
\multirow{4}{*}{SD2.1~\cite{Rombach_2022_CVPR}} & \textit{COS} & \color{ourred}\xmark & 29.0 &23.8&34.2 &  69.8 &67.1 & 72.6 & 60.1&55.7 &66.0 \\
& \textit{L2}& \color{ourred}\xmark &29.1& 23.9&34.3 &  69.8 &66.9 & 72.7 & 61.1& 55.9&66.3 \\
& \textit{COS} & \color{ourgreen}\cmark & 29.6& 24.4&34.9&  70.2 &67.2 & 73.1 & 61.4 &56.2 &66.6 \\
& \textit{L2}& \color{ourgreen}\cmark & \textbf{30.0} & \textbf{24.7}& \textbf{35.2} & \textbf{70.5} & \textbf{67.5}& \textbf{73.5}   & \textbf{61.5}&\textbf{56.2} &\textbf{66.7} \\
\hdashline
\multirow{2}{*}{ADM~\cite{adm}}& \textit{COS}& - & 34.7&29.4&40.1 &  76.1 &\textbf{73.2} & 79.1 & 67.0&62.0 &72.1 \\
& \textit{L2}& - & \textbf{34.9}&\textbf{29.5}&\textbf{40.3} &  \textbf{76.2} &\textbf{73.2} & \textbf{79.2} & \textbf{67.3} &\textbf{62.1} &\textbf{72.5} \\
\bottomrule
\end{NiceTabular}
    \caption{\textbf{ZS-VOS results on multiple benchmarks.}  We compare DINO~\cite{dino}, Stable Diffusion (SD) 2.1~\cite{Rombach_2022_CVPR}, and ADM~\cite{adm} using different similarity functions for affinity, as well as SD with and without prompt learning. \textbf{Bold} denotes the best performing setting for each model.}
    \label{tab:generalization_ability}
\end{table*}

\myparb{Best \mjf\xspace across all diffusion models.} In Fig.~\ref{fig:best_jand_diffmodels}, we show the highest \mjf\xspace for each diffusion model. We observe that ADM~\cite{adm} outperforms all SD variants~\cite{Rombach_2022_CVPR}, suggesting that ImageNet~\cite{russakovsky2015imagenet} pretraining yields better representations for ZS-VOS than LAION~\cite{schuhmann2022laion}. Unless stated otherwise, in the rest of the paper, we will use the layer and time step that yield the highest \mjf\xspace for each diffusion model. 

\myparb{Affinity matrix ablation.} 
Here, we experiment with different similarity functions to compute the affinity matrix for DINO~\cite{dino}, ADM~\cite{adm}, and SD-2.1~\cite{Rombach_2022_CVPR}. In Tab.~\ref{tab:affinity_ablation}, we present our results and observe that the \textit{L1} yields significantly worse results than \textit{COS} and \textit{L2}. Additionally, \textit{L2} improves performance on both DINO~\cite{dino} and ADM~\cite{adm}, yielding a 0.2\% and 0.1\% increase in \mjf, respectively.

\subsection{Correspondences analysis}
\label{exps:corrs_analysis}
We continue our analysis by investigating the correspondences of the affinity matrix (see Sec.~\ref{method:correspondences}). In Fig.~\ref{fig:fg_bg_jandf}, we show the FG-BG percentage in comparison to \mjf\xspace on the DAVIS-17~\cite{davis_17} val set for higher-resolution layers ($\geq 6$) up to 9 and time steps from 0 to 400 across all versions of Stable Diffusion~\cite{Rombach_2022_CVPR}. We observe a strong correlation, indicating that the lower the FG-BG percentage, the higher the \mjf\xspace value. In particular, Spearman's $\rho$ rank correlation is $-0.44$. This finding supports our hypothesis that FG-BG correspondences are harmful for the task ZS-VOS.

\myparb{Filtering out correspondences.} In Tab.~\ref{tab:filter_correspondences}, we show the segmentation quality for all diffusion models using no filtering and our proposed MAG-Filter. We also include an Oracle filter, which uses all ground-truth masks to set the pixels of FG-BG correspondences to 0 in the affinity matrix. We observe that the MAG-Filter yields performance gains ranging from 0.4\% to 0.7\% across all models. When we filter the correspondences by Oracle, we observe a substantial performance gain of 13.5\% in terms of \mjf\xspace, which reveals the crucial impact of correspondences in ZS-VOS.

\myparb{Improving correspondences via prompts.} In Tab.~\ref{tab:prompt_learning}, we show \mjf\xspace when SD-2.1 is prompted with an empty text, the ground-truth text and our prompt learning strategy. The ground-truth text is taken from the caption of the first frame in Ref-DAVIS-17~\cite{khoreva2019video}. We observe a performance boost ranging from 0.1\% to 0.4\% compared to the empty text, indicating the significance of conditioning in SD, as it was also trained with text. An interesting finding is that our prompt learning strategy yields the same \mjf\xspace as when SD is prompted with the ground-truth text, which serves as an oracle since it is not available at test time in ZS-VOS.

\myparb{Cross-Attention maps of Prompt Learning.} In Fig.~\ref{fig:prompt_learning_qualitative}(Left), we show the cross-attention maps ($CA$) learned with the BCE loss for SD-2.1~\cite{Rombach_2022_CVPR}. We observe that the final $CA$ closely aligns with the ground-truth masks. In Fig.~\ref{fig:prompt_learning_qualitative}(Right), we prompt SD with the optimized token for the remaining video frames and observe that $CA$ highlights the object, even though it is optimized using only the first frame. The above findings indicate the effectiveness of our prompt learning strategy, as $CA$ looks temporally coherent. 

\subsection{State-of-the-art zero-shot VOS comparison}
\label{exps:zero_shot_vos}
Tab.~\ref{tab:zero_shot_vos_comp} presents the segmentation performance on the DAVIS-17 val set for state-of-the-art models, alongside their training data. We categorize methods based on whether they are pre-trained on image-level or video-level data and/or fine-tuned on segmentation annotations. We observe that ADM~\cite{adm} with our MAG-Filter, enhanced by our layer and time step findings, outperforms its counterpart, DIFT-ADM~\cite{dift}, by 1.1\% and surpasses all methods that do not use any segmentation annotations and yielding state-of-the-art results.
Among methods trained only on image-level data, Matcher~\cite{liu2023matcher} is the only approach with higher performance than ours, but it clearly benefits from the vast SA-1B~\cite{kirillov2023segment} dataset with 1.1 billion segmentation masks.
This result reveals the strength of diffusion features trained solely on ImageNet~\cite{dosovitskiy2020image}, highlighting their robustness on the ZS-VOS task
despite the lack of direct segmentation supervision, which is labor-intensive and expensive~\cite{coco}.

\subsection{Generalization ability of our findings}
\label{exps:gen_ability}

Here, we demonstrate the generalization of our findings in the previous sections on additional datasets, namely the DAVIS-17~\cite{davis_17} test set and the MOSE~\cite{mose} validation set. 
%
We compare DINO~\cite{dino}, SD 2.1~\cite{Rombach_2022_CVPR}, and ADM~\cite{adm} on Tab.~\ref{tab:generalization_ability}, using \textit{L2} and \textit{COS} similarity for the affinity matrix. For DINO~\cite{dino}, \textit{L2} yields consistent performance boosts across all datasets ranging from 0.1\% to 0.2\% in terms of \mjf. For SD~\cite{Rombach_2022_CVPR}, we also experiment with and without prompt learning. First, we observe that the \textit{L2} distance again yields a performance boost for all datasets, ranging from 0.1\% to 1\%, when comparing results that either both use prompt learning or neither uses it. Prompt learning further improves the segmentation quality, yielding 61.4\% and 61.5\% on the DAVIS-17 test, compared to the SD counterparts using \textit{COS} and \textit{L2} distances without prompt learning, which yield 60.1\% and 61.1\%, respectively. Finally, the above patterns remain consistent with ADM~\cite{adm} outperforming all other models across all datasets. 

\section{Conclusions}
\label{sec:conclusions}
We presented a systematic analysis of Zero-Shot Video Object Segmentation using features from pretrained image diffusion models. We showed that the timestep and layer from which we extract features significantly impact segmentation quality. 
Our findings revealed that point correspondences highly impact performance, highlighting the importance of precise matching in the VOS task.
Diffusion features trained only on ImageNet outperform all other pretrained features on the ZS-VOS task and yield comparable segmentations to models trained on large-scale image segmentation datasets, such as SA-1B~\cite{kirillov2023segment}.

\mypar{Acknowledgements.} V. Kalogeiton was supported by a Hi!Paris collaborative project. D. Papadopoulos was supported by the DFF Sapere Aude Starting Grant ``ACHILLES". We would like to thank O. Kaya and M. Schouten for insightful discussions.

{
    \small
    \bibliographystyle{ieeenat_fullname}
    \bibliography{main}

\begin{thebibliography}{106}
\providecommand{\natexlab}[1]{#1}
\providecommand{\url}[1]{\texttt{#1}}
\expandafter\ifx\csname urlstyle\endcsname\relax
  \providecommand{\doi}[1]{doi: #1}\else
  \providecommand{\doi}{doi: \begingroup \urlstyle{rm}\Url}\fi

\bibitem[Alimohammadi et~al.(2024)Alimohammadi, Nag, Taghanaki, Tagliasacchi, Hamarneh, and Amiri]{alimohammadi2024smite}
Amirhossein Alimohammadi, Sauradip Nag, Saeid~Asgari Taghanaki, Andrea Tagliasacchi, Ghassan Hamarneh, and Ali~Mahdavi Amiri.
\newblock Smite: Segment me in time.
\newblock \emph{arXiv preprint arXiv:2410.18538}, 2024.

\bibitem[Amit et~al.(2021)Amit, Shaharbany, Nachmani, and Wolf]{amit2021segdiff}
Tomer Amit, Tal Shaharbany, Eliya Nachmani, and Lior Wolf.
\newblock Segdiff: Image segmentation with diffusion probabilistic models.
\newblock \emph{arXiv preprint arXiv:2112.00390}, 2021.

\bibitem[Bao et~al.(2018)Bao, Wu, and Liu]{prop_4}
Linchao Bao, Baoyuan Wu, and Wei Liu.
\newblock Cnn in mrf: Video object segmentation via inference in a cnn-based higher-order spatio-temporal mrf.
\newblock In \emph{CVPR}, 2018.

\bibitem[Bekuzarov et~al.(2023)Bekuzarov, Bermudez, Lee, and Li]{bekuzarov2023}
Maksym Bekuzarov, Ariana Bermudez, Joon-Young Lee, and Hao Li.
\newblock Xmem++: Production-level video segmentation from few annotated frames.
\newblock In \emph{ICCV}, 2023.

\bibitem[Caelles et~al.(2017)Caelles, Maninis, Pont-Tuset, Leal-Taix{\'e}, Cremers, and Van~Gool]{on_1}
Sergi Caelles, Kevis-Kokitsi Maninis, Jordi Pont-Tuset, Laura Leal-Taix{\'e}, Daniel Cremers, and Luc Van~Gool.
\newblock One-shot video object segmentation.
\newblock In \emph{CVPR}, 2017.

\bibitem[Carion et~al.(2020)Carion, Massa, Synnaeve, Usunier, Kirillov, and Zagoruyko]{detr}
Nicolas Carion, Francisco Massa, Gabriel Synnaeve, Nicolas Usunier, Alexander Kirillov, and Sergey Zagoruyko.
\newblock End-to-end object detection with transformers.
\newblock In \emph{ECCV}, 2020.

\bibitem[Caron et~al.(2021)Caron, Touvron, Misra, J{\'e}gou, Mairal, Bojanowski, and Joulin]{dino}
Mathilde Caron, Hugo Touvron, Ishan Misra, Herv{\'e} J{\'e}gou, Julien Mairal, Piotr Bojanowski, and Armand Joulin.
\newblock Emerging properties in self-supervised vision transformers.
\newblock In \emph{ICCV}, 2021.

\bibitem[Chen et~al.(2023{\natexlab{a}})Chen, Sun, Song, and Luo]{Chen_2023_ICCV}
Shoufa Chen, Peize Sun, Yibing Song, and Ping Luo.
\newblock Diffusiondet: Diffusion model for object detection.
\newblock In \emph{ICCV}, 2023{\natexlab{a}}.

\bibitem[Chen et~al.(2023{\natexlab{b}})Chen, Li, Saxena, Hinton, and Fleet]{chen2023generalist}
Ting Chen, Lala Li, Saurabh Saxena, Geoffrey Hinton, and David~J Fleet.
\newblock A generalist framework for panoptic segmentation of images and videos.
\newblock In \emph{ICCV}, 2023{\natexlab{b}}.

\bibitem[Chen et~al.(2020)Chen, Li, Yuan, Yu, Shen, and Qi]{prop_6}
Xi Chen, Zuoxin Li, Ye Yuan, Gang Yu, Jianxin Shen, and Donglian Qi.
\newblock State-aware tracker for real-time video object segmentation.
\newblock In \emph{CVPR}, 2020.

\bibitem[Cheng and Schwing(2022)]{xmem}
Ho~Kei Cheng and Alexander~G Schwing.
\newblock Xmem: Long-term video object segmentation with an atkinson-shiffrin memory model.
\newblock In \emph{ECCV}, 2022.

\bibitem[Cheng et~al.(2020)Cheng, Chung, Tai, and Tang]{cheng2020cascadepsp}
Ho~Kei Cheng, Jihoon Chung, Yu-Wing Tai, and Chi-Keung Tang.
\newblock Cascadepsp: Toward class-agnostic and very high-resolution segmentation via global and local refinement.
\newblock In \emph{CVPR}, 2020.

\bibitem[Cheng et~al.(2021)Cheng, Tai, and Tang]{stcn}
Ho~Kei Cheng, Yu-Wing Tai, and Chi-Keung Tang.
\newblock Rethinking space-time networks with improved memory coverage for efficient video object segmentation.
\newblock In \emph{NeurIPS}, 2021.

\bibitem[Cheng et~al.(2024)Cheng, Oh, Price, Lee, and Schwing]{cutie}
Ho~Kei Cheng, Seoung~Wug Oh, Brian Price, Joon-Young Lee, and Alexander Schwing.
\newblock Putting the object back into video object segmentation.
\newblock In \emph{CVPR}, 2024.

\bibitem[Cheng et~al.(2018)Cheng, Tsai, Hung, Wang, and Yang]{prop_5}
Jingchun Cheng, Yi-Hsuan Tsai, Wei-Chih Hung, Shengjin Wang, and Ming-Hsuan Yang.
\newblock Fast and accurate online video object segmentation via tracking parts.
\newblock In \emph{CVPR}, 2018.

\bibitem[Cheng et~al.(2015)Cheng, Mitra, Huang, Torr, and Hu]{msr30k}
Ming-Ming Cheng, Niloy~J. Mitra, Xiaolei Huang, Philip H.~S. Torr, and Shi-Min Hu.
\newblock Global contrast based salient region detection.
\newblock \emph{IEEE Transactions on Pattern Analysis and Machine Intelligence}, 2015.

\bibitem[Delatolas et~al.(2023)Delatolas, Kalogeiton, and Papadopoulos]{delatolas2023eva}
Thanos Delatolas, Vicky Kalogeiton, and Dim~P Papadopoulos.
\newblock Eva-vos: Efficient video annotation for video object segmentation.
\newblock In \emph{ICCVW CVEU}, 2023.

\bibitem[Delatolas et~al.(2024)Delatolas, Kalogeiton, and Papadopoulos]{delatolas2024learning}
Thanos Delatolas, Vicky Kalogeiton, and Dim~P. Papadopoulos.
\newblock Learning the what and how of annotation in video object segmentation.
\newblock In \emph{WACV}, 2024.

\bibitem[Dhariwal and Nichol(2021)]{adm}
Prafulla Dhariwal and Alexander Nichol.
\newblock Diffusion models beat gans on image synthesis.
\newblock \emph{Advances in neural information processing systems}, 34, 2021.

\bibitem[Ding et~al.(2023)Ding, Liu, He, Jiang, Torr, and Bai]{mose}
Henghui Ding, Chang Liu, Shuting He, Xudong Jiang, Philip~HS Torr, and Song Bai.
\newblock {MOSE}: A new dataset for video object segmentation in complex scenes.
\newblock In \emph{ICCV}, 2023.

\bibitem[Dosovitskiy et~al.(2020{\natexlab{a}})Dosovitskiy, Beyer, Kolesnikov, Weissenborn, Zhai, Unterthiner, Dehghani, Minderer, Heigold, Gelly, et~al.]{dosovitskiy2020image}
Alexey Dosovitskiy, Lucas Beyer, Alexander Kolesnikov, Dirk Weissenborn, Xiaohua Zhai, Thomas Unterthiner, Mostafa Dehghani, Matthias Minderer, Georg Heigold, Sylvain Gelly, et~al.
\newblock An image is worth 16x16 words: Transformers for image recognition at scale.
\newblock \emph{arXiv preprint arXiv:2010.11929}, 2020{\natexlab{a}}.

\bibitem[Dosovitskiy et~al.(2020{\natexlab{b}})Dosovitskiy, Beyer, Kolesnikov, Weissenborn, Zhai, Unterthiner, Dehghani, Minderer, Heigold, Gelly, et~al.]{vit}
Alexey Dosovitskiy, Lucas Beyer, Alexander Kolesnikov, Dirk Weissenborn, Xiaohua Zhai, Thomas Unterthiner, Mostafa Dehghani, Matthias Minderer, Georg Heigold, Sylvain Gelly, et~al.
\newblock An image is worth 16x16 words: Transformers for image recognition at scale.
\newblock \emph{arXiv preprint arXiv:2010.11929}, 2020{\natexlab{b}}.

\bibitem[Esser et~al.(2021)Esser, Rombach, and Ommer]{Esser_2021_CVPR}
Patrick Esser, Robin Rombach, and Bjorn Ommer.
\newblock Taming transformers for high-resolution image synthesis.
\newblock In \emph{CVPR}, 2021.

\bibitem[Esser et~al.(2023)Esser, Chiu, Atighehchian, Granskog, and Germanidis]{esser2023structure}
Patrick Esser, Johnathan Chiu, Parmida Atighehchian, Jonathan Granskog, and Anastasis Germanidis.
\newblock Structure and content-guided video synthesis with diffusion models.
\newblock In \emph{Proceedings of the IEEE/CVF international conference on computer vision}, 2023.

\bibitem[Esser et~al.(2024)Esser, Kulal, Blattmann, Entezari, M{\"u}ller, Saini, Levi, Lorenz, Sauer, Boesel, et~al.]{stable_diffusion_3}
Patrick Esser, Sumith Kulal, Andreas Blattmann, Rahim Entezari, Jonas M{\"u}ller, Harry Saini, Yam Levi, Dominik Lorenz, Axel Sauer, Frederic Boesel, et~al.
\newblock Scaling rectified flow transformers for high-resolution image synthesis.
\newblock In \emph{ICML}, 2024.

\bibitem[Everingham et~al.(2015)Everingham, Eslami, Van~Gool, Williams, Winn, and Zisserman]{voc}
M. Everingham, S.~M.~A. Eslami, L. Van~Gool, C.~K.~I. Williams, J. Winn, and A. Zisserman.
\newblock The pascal visual object classes challenge: A retrospective.
\newblock \emph{IJCV}, 2015.

\bibitem[Fragkiadaki et~al.(2015)Fragkiadaki, Arbelaez, Felsen, and Malik]{fragkiadaki2015learning}
Katerina Fragkiadaki, Pablo Arbelaez, Panna Felsen, and Jitendra Malik.
\newblock Learning to segment moving objects in videos.
\newblock In \emph{CVPR}, 2015.

\bibitem[Gong et~al.(2023)Gong, Danelljan, Sun, Mangas, and Van~Gool]{gong2023prompting}
Rui Gong, Martin Danelljan, Han Sun, Julio~Delgado Mangas, and Luc Van~Gool.
\newblock Prompting diffusion representations for cross-domain semantic segmentation.
\newblock \emph{arXiv preprint arXiv:2307.02138}, 2023.

\bibitem[Goyal et~al.(2024)Goyal, Fan, Siam, and Sigal]{goyal2024tamvt}
Raghav Goyal, Wan-Cyuan Fan, Mennatullah Siam, and Leonid Sigal.
\newblock Tam-vt: Transformation-aware multi-scale video transformer for segmentation and tracking.
\newblock \emph{arXiv preprint arXiv:2312.08514}, 2024.

\bibitem[Gupta et~al.(2019)Gupta, Dollar, and Girshick]{gupta2019lvis}
Agrim Gupta, Piotr Dollar, and Ross Girshick.
\newblock Lvis: A dataset for large vocabulary instance segmentation.
\newblock In \emph{CVPR}, 2019.

\bibitem[He et~al.(2016)He, Zhang, Ren, and Sun]{he2016deep}
Kaiming He, Xiangyu Zhang, Shaoqing Ren, and Jian Sun.
\newblock Deep residual learning for image recognition.
\newblock In \emph{CVPR}, 2016.

\bibitem[He et~al.(2019)He, Fan, Wu, Xie, and Girshick]{he2019moco}
Kaiming He, Haoqi Fan, Yuxin Wu, Saining Xie, and Ross Girshick.
\newblock Momentum contrast for unsupervised visual representation learning.
\newblock \emph{arXiv preprint arXiv:1911.05722}, 2019.

\bibitem[Ho and Salimans(2022)]{ho2022classifier}
Jonathan Ho and Tim Salimans.
\newblock Classifier-free diffusion guidance.
\newblock \emph{arXiv preprint arXiv:2207.12598}, 2022.

\bibitem[Ho et~al.(2020)Ho, Jain, and Abbeel]{ddpm}
Jonathan Ho, Ajay Jain, and Pieter Abbeel.
\newblock Denoising diffusion probabilistic models.
\newblock \emph{Advances in neural information processing systems}, 2020.

\bibitem[Jabri et~al.(2020)Jabri, Owens, and Efros]{jabri2020space}
Allan Jabri, Andrew Owens, and Alexei Efros.
\newblock Space-time correspondence as a contrastive random walk.
\newblock \emph{NeurIPS}, 2020.

\bibitem[Jampani et~al.(2017)Jampani, Gadde, and Gehler]{prop_3}
Varun Jampani, Raghudeep Gadde, and Peter~V Gehler.
\newblock Video propagation networks.
\newblock In \emph{CVPR}, 2017.

\bibitem[Jang and Kim(2017)]{prop_2}
Won-Dong Jang and Chang-Su Kim.
\newblock Online video object segmentation via convolutional trident network.
\newblock In \emph{CVPR}, 2017.

\bibitem[Kay et~al.(2017)Kay, Carreira, Simonyan, Zhang, Hillier, Vijayanarasimhan, Viola, Green, Back, Natsev, et~al.]{kay2017kinetics}
Will Kay, Joao Carreira, Karen Simonyan, Brian Zhang, Chloe Hillier, Sudheendra Vijayanarasimhan, Fabio Viola, Tim Green, Trevor Back, Paul Natsev, et~al.
\newblock The kinetics human action video dataset.
\newblock \emph{arXiv preprint arXiv:1705.06950}, 2017.

\bibitem[Ke et~al.(2024)Ke, Obukhov, Huang, Metzger, Daudt, and Schindler]{ke2023repurposing}
Bingxin Ke, Anton Obukhov, Shengyu Huang, Nando Metzger, Rodrigo~Caye Daudt, and Konrad Schindler.
\newblock Repurposing diffusion-based image generators for monocular depth estimation.
\newblock In \emph{CVPR}, 2024.

\bibitem[Khani et~al.(2023)Khani, Taghanaki, Sanghi, Amiri, and Hamarneh]{khani2023slime}
Aliasghar Khani, Saeid~Asgari Taghanaki, Aditya Sanghi, Ali~Mahdavi Amiri, and Ghassan Hamarneh.
\newblock Slime: Segment like me.
\newblock \emph{arXiv preprint arXiv:2309.03179}, 2023.

\bibitem[Khoreva et~al.(2019)Khoreva, Rohrbach, and Schiele]{khoreva2019video}
Anna Khoreva, Anna Rohrbach, and Bernt Schiele.
\newblock Video object segmentation with language referring expressions.
\newblock In \emph{ACCV}, 2019.

\bibitem[Kirillov et~al.(2023)Kirillov, Mintun, Ravi, Mao, Rolland, Gustafson, Xiao, Whitehead, Berg, Lo, et~al.]{kirillov2023segment}
Alexander Kirillov, Eric Mintun, Nikhila Ravi, Hanzi Mao, Chloe Rolland, Laura Gustafson, Tete Xiao, Spencer Whitehead, Alexander~C Berg, Wan-Yen Lo, et~al.
\newblock Segment anything.
\newblock In \emph{Proceedings of the IEEE/CVF international conference on computer vision}, pages 4015--4026, 2023.

\bibitem[Kondapaneni et~al.(2024)Kondapaneni, Marks, Knott, Guimaraes, and Perona]{kondapaneni2024tadp}
Neehar Kondapaneni, Markus Marks, Manuel Knott, Rogerio Guimaraes, and Pietro Perona.
\newblock Text-image alignment for diffusion-based perception.
\newblock \emph{CVPR}, 2024.

\bibitem[Li et~al.(2023{\natexlab{a}})Li, Wang, Zhou, Li, and Yang]{li2023unified}
Liulei Li, Wenguan Wang, Tianfei Zhou, Jianwu Li, and Yi Yang.
\newblock Unified mask embedding and correspondence learning for self-supervised video segmentation.
\newblock In \emph{CVPR}, 2023{\natexlab{a}}.

\bibitem[Li and Liu(2023)]{stt}
Rui Li and Dong Liu.
\newblock Spatial-then-temporal self-supervised learning for video correspondence.
\newblock In \emph{CVPR}, 2023.

\bibitem[Li et~al.(2023{\natexlab{b}})Li, Zhou, and Liu]{li2023learning}
Rui Li, Shenglong Zhou, and Dong Liu.
\newblock Learning fine-grained features for pixel-wise video correspondences.
\newblock In \emph{ICCV}, 2023{\natexlab{b}}.

\bibitem[Li et~al.(2020)Li, Wei, Chen, Tai, and Tang]{li2020fss}
Xiang Li, Tianhan Wei, Yau~Pun Chen, Yu-Wing Tai, and Chi-Keung Tang.
\newblock Fss-1000: A 1000-class dataset for few-shot segmentation.
\newblock In \emph{CVPR}, 2020.

\bibitem[Li et~al.(2024)Li, Miao, He, Wang, Lu, and Yang]{li2024learning}
Xin Li, Deshui Miao, Zhenyu He, Yaowei Wang, Huchuan Lu, and Ming-Hsuan Yang.
\newblock Learning spatial-semantic features for robust video object segmentation.
\newblock In \emph{ICLR}, 2024.

\bibitem[Lin et~al.(2014)Lin, Maire, Belongie, Hays, Perona, Ramanan, Doll{\'a}r, and Zitnick]{coco}
Tsung-Yi Lin, Michael Maire, Serge Belongie, James Hays, Pietro Perona, Deva Ramanan, Piotr Doll{\'a}r, and C~Lawrence Zitnick.
\newblock Microsoft coco: Common objects in context.
\newblock In \emph{ECCV}, 2014.

\bibitem[Lipman et~al.(2022)Lipman, Chen, Ben-Hamu, Nickel, and Le]{lipman2022flow}
Yaron Lipman, Ricky~TQ Chen, Heli Ben-Hamu, Maximilian Nickel, and Matt Le.
\newblock Flow matching for generative modeling.
\newblock \emph{arXiv preprint arXiv:2210.02747}, 2022.

\bibitem[Liu et~al.(2023)Liu, Zhu, Li, Chen, Wang, and Shen]{liu2023matcher}
Yang Liu, Muzhi Zhu, Hengtao Li, Hao Chen, Xinlong Wang, and Chunhua Shen.
\newblock Matcher: Segment anything with one shot using all-purpose feature matching.
\newblock \emph{arXiv preprint arXiv:2305.13310}, 2023.

\bibitem[Maninis et~al.(2018)Maninis, Caelles, Chen, Pont-Tuset, Leal-Taix{\'e}, Cremers, and Van~Gool]{on_5}
K-K Maninis, Sergi Caelles, Yuhua Chen, Jordi Pont-Tuset, Laura Leal-Taix{\'e}, Daniel Cremers, and Luc Van~Gool.
\newblock Video object segmentation without temporal information.
\newblock \emph{PAMI}, 2018.

\bibitem[Mayer et~al.(2016)Mayer, Ilg, Hausser, Fischer, Cremers, Dosovitskiy, and Brox]{Mayer_2016_CVPR}
Nikolaus Mayer, Eddy Ilg, Philip Hausser, Philipp Fischer, Daniel Cremers, Alexey Dosovitskiy, and Thomas Brox.
\newblock A large dataset to train convolutional networks for disparity, optical flow, and scene flow estimation.
\newblock In \emph{PCVPR}, 2016.

\bibitem[Meinhardt and Leal-Taix{\'e}(2020)]{on_6}
Tim Meinhardt and Laura Leal-Taix{\'e}.
\newblock Make one-shot video object segmentation efficient again.
\newblock \emph{NeurIPS}, 2020.

\bibitem[Meng et~al.(2024)Meng, Lan, Li, Alvarez, Wu, and Jiang]{meng2024segic}
Lingchen Meng, Shiyi Lan, Hengduo Li, Jose~M Alvarez, Zuxuan Wu, and Yu-Gang Jiang.
\newblock Segic: Unleashing the emergent correspondence for in-context segmentation.
\newblock In \emph{ECCV}, 2024.

\bibitem[Namekata et~al.(2024)Namekata, Sabour, Fidler, and Kim]{namekata2024emerdiff}
Koichi Namekata, Amirmojtaba Sabour, Sanja Fidler, and Seung~Wook Kim.
\newblock Emerdiff: Emerging pixel-level semantic knowledge in diffusion models.
\newblock \emph{arXiv preprint arXiv:2401.11739}, 2024.

\bibitem[Oh et~al.(2019)Oh, Lee, Xu, and Kim]{stm}
Seoung~Wug Oh, Joon-Young Lee, Ning Xu, and Seon~Joo Kim.
\newblock Video object segmentation using space-time memory networks.
\newblock In \emph{ICCV}, 2019.

\bibitem[Pan et~al.(2022)Pan, Li, Yang, Zhou, Zhou, Yang, Zhou, and Yang]{invos}
Xiao Pan, Peike Li, Zongxin Yang, Huiling Zhou, Chang Zhou, Hongxia Yang, Jingren Zhou, and Yi Yang.
\newblock In-n-out generative learning for dense unsupervised video segmentation.
\newblock In \emph{ACM}, 2022.

\bibitem[Peebles and Xie(2023)]{dit}
William Peebles and Saining Xie.
\newblock Scalable diffusion models with transformers.
\newblock In \emph{ICCV}, 2023.

\bibitem[Peng et~al.(2024)Peng, Zhang, Hu, Ke, Yau, and Liu]{peng2024harnessing}
Duo Peng, Zhengbo Zhang, Ping Hu, Qiuhong Ke, David~KY Yau, and Jun Liu.
\newblock Harnessing text-to-image diffusion models for category-agnostic pose estimation.
\newblock In \emph{ECCV}. Springer, 2024.

\bibitem[Perazzi et~al.(2016)Perazzi, Pont-Tuset, McWilliams, Van~Gool, Gross, and Sorkine-Hornung]{perazzi2016benchmark}
Federico Perazzi, Jordi Pont-Tuset, Brian McWilliams, Luc Van~Gool, Markus Gross, and Alexander Sorkine-Hornung.
\newblock A benchmark dataset and evaluation methodology for video object segmentation.
\newblock In \emph{CVPR}, 2016.

\bibitem[Perazzi et~al.(2017)Perazzi, Khoreva, Benenson, Schiele, and Sorkine-Hornung]{prop_1}
Federico Perazzi, Anna Khoreva, Rodrigo Benenson, Bernt Schiele, and Alexander Sorkine-Hornung.
\newblock Learning video object segmentation from static images.
\newblock In \emph{CVPR}, 2017.

\bibitem[Pnvr et~al.(2023)Pnvr, Singh, Ghosh, Siddiquie, and Jacobs]{ldznet}
Koutilya Pnvr, Bharat Singh, Pallabi Ghosh, Behjat Siddiquie, and David Jacobs.
\newblock Ld-znet: A latent diffusion approach for text-based image segmentation.
\newblock In \emph{ICCV}, 2023.

\bibitem[Pont-Tuset et~al.(2017)Pont-Tuset, Perazzi, Caelles, Arbel{\'a}ez, Sorkine-Hornung, and Van~Gool]{davis_17}
Jordi Pont-Tuset, Federico Perazzi, Sergi Caelles, Pablo Arbel{\'a}ez, Alex Sorkine-Hornung, and Luc Van~Gool.
\newblock The 2017 davis challenge on video object segmentation.
\newblock \emph{arXiv preprint arXiv:1704.00675}, 2017.

\bibitem[Radford et~al.(2021)Radford, Kim, Hallacy, Ramesh, Goh, Agarwal, Sastry, Askell, Mishkin, Clark, et~al.]{clip}
Alec Radford, Jong~Wook Kim, Chris Hallacy, Aditya Ramesh, Gabriel Goh, Sandhini Agarwal, Girish Sastry, Amanda Askell, Pamela Mishkin, Jack Clark, et~al.
\newblock Learning transferable visual models from natural language supervision.
\newblock In \emph{ICML}, pages 8748--8763, 2021.

\bibitem[Ravi et~al.(2024)Ravi, Gabeur, Hu, Hu, Ryali, Ma, Khedr, R{\"a}dle, Rolland, Gustafson, et~al.]{ravi2024sam}
Nikhila Ravi, Valentin Gabeur, Yuan-Ting Hu, Ronghang Hu, Chaitanya Ryali, Tengyu Ma, Haitham Khedr, Roman R{\"a}dle, Chloe Rolland, Laura Gustafson, et~al.
\newblock Sam 2: Segment anything in images and videos.
\newblock \emph{arXiv preprint arXiv:2408.00714}, 2024.

\bibitem[Rombach et~al.(2022)Rombach, Blattmann, Lorenz, Esser, and Ommer]{Rombach_2022_CVPR}
Robin Rombach, Andreas Blattmann, Dominik Lorenz, Patrick Esser, and Bj\"orn Ommer.
\newblock High-resolution image synthesis with latent diffusion models.
\newblock In \emph{CVPR}, 2022.

\bibitem[Ronneberger et~al.(2015)Ronneberger, Fischer, and Brox]{ronneberger2015u}
Olaf Ronneberger, Philipp Fischer, and Thomas Brox.
\newblock U-net: Convolutional networks for biomedical image segmentation.
\newblock In \emph{MICCAI}, 2015.

\bibitem[Russakovsky et~al.(2015)Russakovsky, Deng, Su, Krause, Satheesh, Ma, Huang, Karpathy, Khosla, Bernstein, et~al.]{russakovsky2015imagenet}
Olga Russakovsky, Jia Deng, Hao Su, Jonathan Krause, Sanjeev Satheesh, Sean Ma, Zhiheng Huang, Andrej Karpathy, Aditya Khosla, Michael Bernstein, et~al.
\newblock Imagenet large scale visual recognition challenge.
\newblock \emph{International journal of computer vision}, 2015.

\bibitem[Santos et~al.(2023)Santos, da~Silva, and Oliveira]{Santos_2023_BMVC}
Marcelo~M Santos, Jefferson~Fontinele da Silva, and Luciano Oliveira.
\newblock Shls: Superfeatures learned from still images for self-supervised vos.
\newblock In \emph{BMVC}, 2023.

\bibitem[Saxena et~al.(2023)Saxena, Herrmann, Hur, Kar, Norouzi, Sun, and Fleet]{saxena2023surprising}
Saurabh Saxena, Charles Herrmann, Junhwa Hur, Abhishek Kar, Mohammad Norouzi, Deqing Sun, and David~J Fleet.
\newblock The surprising effectiveness of diffusion models for optical flow and monocular depth estimation.
\newblock \emph{NeurIPS}, 2023.

\bibitem[Schuhmann et~al.(2022)Schuhmann, Beaumont, Vencu, Gordon, Wightman, Cherti, Coombes, Katta, Mullis, Wortsman, et~al.]{schuhmann2022laion}
Christoph Schuhmann, Romain Beaumont, Richard Vencu, Cade Gordon, Ross Wightman, Mehdi Cherti, Theo Coombes, Aarush Katta, Clayton Mullis, Mitchell Wortsman, et~al.
\newblock Laion-5b: An open large-scale dataset for training next generation image-text models.
\newblock \emph{NeurIPS}, 2022.

\bibitem[Shi et~al.(2015)Shi, Yan, Xu, and Jia]{shi2015hierarchicalECSSD}
Jianping Shi, Qiong Yan, Li Xu, and Jiaya Jia.
\newblock Hierarchical image saliency detection on extended cssd.
\newblock In \emph{TPAMI}, 2015.

\bibitem[Song et~al.(2020)Song, Sohl-Dickstein, Kingma, Kumar, Ermon, and Poole]{song2020score}
Yang Song, Jascha Sohl-Dickstein, Diederik~P Kingma, Abhishek Kumar, Stefano Ermon, and Ben Poole.
\newblock Score-based generative modeling through stochastic differential equations.
\newblock \emph{arXiv preprint arXiv:2011.13456}, 2020.

\bibitem[Stracke et~al.(2024)Stracke, Baumann, Bauer, Fundel, and Ommer]{stracke2024cleandift}
Nick Stracke, Stefan~Andreas Baumann, Kolja Bauer, Frank Fundel, and Bj{\"o}rn Ommer.
\newblock Cleandift: Diffusion features without noise.
\newblock \emph{arXiv preprint arXiv:2412.03439}, 2024.

\bibitem[Tang et~al.(2023)Tang, Jia, Wang, Phoo, and Hariharan]{dift}
Luming Tang, Menglin Jia, Qianqian Wang, Cheng~Perng Phoo, and Bharath Hariharan.
\newblock Emergent correspondence from image diffusion.
\newblock In \emph{Advances in Neural Information Processing Systems}, 2023.

\bibitem[Tokmakov et~al.(2022)Tokmakov, Li, and Gaidon]{vost}
Pavel Tokmakov, Jie Li, and Adrien Gaidon.
\newblock Breaking the" object" in video object segmentation.
\newblock \emph{arXiv preprint arXiv:2212.06200}, 2022.

\bibitem[Uziel et~al.(2023)Uziel, Dinari, and Freifeld]{uziel2023vit}
Roy Uziel, Or Dinari, and Oren Freifeld.
\newblock From vit features to training-free video object segmentation via streaming-data mixture models.
\newblock In \emph{NeurIPS}, 2023.

\bibitem[Ventura et~al.(2019)Ventura, Bellver, Girbau, Salvador, Marques, and Giro-i Nieto]{prop_7}
Carles Ventura, Miriam Bellver, Andreu Girbau, Amaia Salvador, Ferran Marques, and Xavier Giro-i Nieto.
\newblock Rvos: End-to-end recurrent network for video object segmentation.
\newblock In \emph{CVPR}, 2019.

\bibitem[Voigtlaender and Leibe(2017)]{on_2}
Paul Voigtlaender and Bastian Leibe.
\newblock Online adaptation of convolutional neural networks for video object segmentation.
\newblock \emph{BMCV}, 2017.

\bibitem[Wang et~al.(2023{\natexlab{a}})Wang, Chen, Wu, Luo, Tang, Dai, Zhao, Xie, Yuan, and Jiang]{Wang_2023_CVPR}
Junke Wang, Dongdong Chen, Zuxuan Wu, Chong Luo, Chuanxin Tang, Xiyang Dai, Yucheng Zhao, Yujia Xie, Lu Yuan, and Yu-Gang Jiang.
\newblock Look before you match: Instance understanding matters in video object segmentation.
\newblock In \emph{CVPR}, 2023{\natexlab{a}}.

\bibitem[Wang et~al.(2017)Wang, Lu, Wang, Feng, Wang, Yin, and Ruan]{wang2017learning}
Lijun Wang, Huchuan Lu, Yifan Wang, Mengyang Feng, Dong Wang, Baocai Yin, and Xiang Ruan.
\newblock Learning to detect salient objects with image-level supervision.
\newblock In \emph{CVPR}, 2017.

\bibitem[Wang et~al.(2023{\natexlab{b}})Wang, Wang, Cao, Shen, and Huang]{wang2023images}
Xinlong Wang, Wen Wang, Yue Cao, Chunhua Shen, and Tiejun Huang.
\newblock Images speak in images: A generalist painter for in-context visual learning.
\newblock In \emph{Proceedings of the IEEE/CVF Conference on Computer Vision and Pattern Recognition}, pages 6830--6839, 2023{\natexlab{b}}.

\bibitem[Wang et~al.(2023{\natexlab{c}})Wang, Zhang, Cao, Wang, Shen, and Huang]{wang2023seggpt}
Xinlong Wang, Xiaosong Zhang, Yue Cao, Wen Wang, Chunhua Shen, and Tiejun Huang.
\newblock Seggpt: Segmenting everything in context.
\newblock \emph{arXiv preprint arXiv:2304.03284}, 2023{\natexlab{c}}.

\bibitem[Wu et~al.(2023{\natexlab{a}})Wu, Yang, Wu, and Chan]{Wu_2023_ICCV_VOS}
Qiangqiang Wu, Tianyu Yang, Wei Wu, and Antoni~B. Chan.
\newblock Scalable video object segmentation with simplified framework.
\newblock In \emph{ICCV}, 2023{\natexlab{a}}.

\bibitem[Wu et~al.()Wu, Zhao, Shou, Zhou, and Shen]{Wu_2023_ICCV}
Weijia Wu, Yuzhong Zhao, Mike~Zheng Shou, Hong Zhou, and Chunhua Shen.
\newblock Diffumask: Synthesizing images with pixel-level annotations for semantic segmentation using diffusion models.
\newblock In \emph{ICCV}.

\bibitem[Wu et~al.(2023{\natexlab{b}})Wu, Zhao, Chen, Gu, Zhao, He, Zhou, Shou, and Shen]{wu2023datasetdm}
Weijia Wu, Yuzhong Zhao, Hao Chen, Yuchao Gu, Rui Zhao, Yefei He, Hong Zhou, Mike~Zheng Shou, and Chunhua Shen.
\newblock Datasetdm: Synthesizing data with perception annotations using diffusion models.
\newblock \emph{NeurIPS}, 2023{\natexlab{b}}.

\bibitem[Xiao et~al.(2018)Xiao, Feng, Lin, Liu, and Zhang]{on_4}
Huaxin Xiao, Jiashi Feng, Guosheng Lin, Yu Liu, and Maojun Zhang.
\newblock Monet: Deep motion exploitation for video object segmentation.
\newblock In \emph{CVPR}, 2018.

\bibitem[Xie et~al.(2021)Xie, Yao, Zhou, Zhang, and Sun]{xie2021efficient}
Haozhe Xie, Hongxun Yao, Shangchen Zhou, Shengping Zhang, and Wenxiu Sun.
\newblock Efficient regional memory network for video object segmentation.
\newblock In \emph{CVPR}, 2021.

\bibitem[Xu et~al.(2023)Xu, Liu, Vahdat, Byeon, Wang, and De~Mello]{xu2023odise}
Jiarui Xu, Sifei Liu, Arash Vahdat, Wonmin Byeon, Xiaolong Wang, and Shalini De~Mello.
\newblock {Open-Vocabulary Panoptic Segmentation with Text-to-Image Diffusion Models}.
\newblock \emph{arXiv preprint arXiv:2303.04803}, 2023.

\bibitem[Xu et~al.(2018)Xu, Yang, Fan, Yang, Yue, Liang, Price, Cohen, and Huang]{youtube_vos}
Ning Xu, Linjie Yang, Yuchen Fan, Jianchao Yang, Dingcheng Yue, Yuchen Liang, Brian Price, Scott Cohen, and Thomas Huang.
\newblock Youtube-vos: Sequence-to-sequence video object segmentation.
\newblock In \emph{ECCV}, 2018.

\bibitem[Yang et~al.(2018)Yang, Wang, Xiong, Yang, and Katsaggelos]{on_3}
Linjie Yang, Yanran Wang, Xuehan Xiong, Jianchao Yang, and Aggelos~K Katsaggelos.
\newblock Efficient video object segmentation via network modulation.
\newblock In \emph{CVPR}, 2018.

\bibitem[Yang and Yang(2022)]{deaot}
Zongxin Yang and Yi Yang.
\newblock Decoupling features in hierarchical propagation for video object segmentation.
\newblock \emph{NeurIPS}, 2022.

\bibitem[Yang et~al.(2021)Yang, Wei, and Yang]{aot}
Zongxin Yang, Yunchao Wei, and Yi Yang.
\newblock Associating objects with transformers for video object segmentation.
\newblock \emph{NeurIPS}, 34, 2021.

\bibitem[Zeng et~al.(2019)Zeng, Zhang, Zhang, Lin, and Lu]{zeng2019towardsHRSOD}
Yi Zeng, Pingping Zhang, Jianming Zhang, Zhe Lin, and Huchuan Lu.
\newblock Towards high-resolution salient object detection.
\newblock In \emph{ICCV}, 2019.

\bibitem[Zhang et~al.({\natexlab{a}})Zhang, Herrmann, Hur, Chen, Jampani, Sun, and Yang]{Zhang_2024_CVPR}
Junyi Zhang, Charles Herrmann, Junhwa Hur, Eric Chen, Varun Jampani, Deqing Sun, and Ming-Hsuan Yang.
\newblock Telling left from right: Identifying geometry-aware semantic correspondence.
\newblock In \emph{CVPR}, {\natexlab{a}}.

\bibitem[Zhang et~al.(2023{\natexlab{a}})Zhang, Herrmann, Hur, Polania~Cabrera, Jampani, Sun, and Yang]{zhang2023tale}
Junyi Zhang, Charles Herrmann, Junhwa Hur, Luisa Polania~Cabrera, Varun Jampani, Deqing Sun, and Ming-Hsuan Yang.
\newblock A tale of two features: Stable diffusion complements dino for zero-shot semantic correspondence.
\newblock \emph{NeurIPS}, 2023{\natexlab{a}}.

\bibitem[Zhang et~al.({\natexlab{b}})Zhang, Song, Shi, Liu, and Li]{zhang2024three}
Manyuan Zhang, Guanglu Song, Xiaoyu Shi, Yu Liu, and Hongsheng Li.
\newblock Three things we need to know about transferring stable diffusion to visual dense prediction tasks.
\newblock In \emph{ECCV}. Springer, {\natexlab{b}}.

\bibitem[Zhang et~al.(2023{\natexlab{b}})Zhang, Jiang, Guo, Yan, Pan, Dong, Gao, and Li]{zhang2023personalize}
Renrui Zhang, Zhengkai Jiang, Ziyu Guo, Shilin Yan, Junting Pan, Hao Dong, Peng Gao, and Hongsheng Li.
\newblock Personalize segment anything model with one shot.
\newblock \emph{arXiv preprint arXiv:2305.03048}, 2023{\natexlab{b}}.

\bibitem[Zhang et~al.(2024)Zhang, Xu, Peng, Rahmani, and Liu]{zhang2024diff}
Zhengbo Zhang, Li Xu, Duo Peng, Hossein Rahmani, and Jun Liu.
\newblock Diff-tracker: text-to-image diffusion models are unsupervised trackers.
\newblock In \emph{ECCV}. Springer, 2024.

\bibitem[Zhao et~al.(2023)Zhao, Rao, Liu, Liu, Zhou, and Lu]{vpd}
Wenliang Zhao, Yongming Rao, Zuyan Liu, Benlin Liu, Jie Zhou, and Jiwen Lu.
\newblock Unleashing text-to-image diffusion models for visual perception.
\newblock In \emph{ICCV}, 2023.

\bibitem[Zhou et~al.(2017)Zhou, Zhao, Puig, Fidler, Barriuso, and Torralba]{ade20k_1}
Bolei Zhou, Hang Zhao, Xavier Puig, Sanja Fidler, Adela Barriuso, and Antonio Torralba.
\newblock Scene parsing through ade20k dataset.
\newblock In \emph{CVPR}, 2017.

\bibitem[Zhou et~al.(2019)Zhou, Zhao, Puig, Xiao, Fidler, Barriuso, and Torralba]{ade20k_2}
Bolei Zhou, Hang Zhao, Xavier Puig, Tete Xiao, Sanja Fidler, Adela Barriuso, and Antonio Torralba.
\newblock Semantic understanding of scenes through the ade20k dataset.
\newblock \emph{IJCV}, 2019.

\bibitem[Zhou et~al.(2024)Zhou, Pang, and Wang]{zhou2024rmem}
Junbao Zhou, Ziqi Pang, and Yu-Xiong Wang.
\newblock Rmem: Restricted memory banks improve video object segmentation.
\newblock In \emph{CVPR}, 2024.

\bibitem[Zhu et~al.(2024)Zhu, Feng, Chen, Yuan, Qiao, and Hua]{zhu2024exploring}
Zixin Zhu, Xuelu Feng, Dongdong Chen, Junsong Yuan, Chunming Qiao, and Gang Hua.
\newblock Exploring pre-trained text-to-video diffusion models for referring video object segmentation.
\newblock In \emph{ECCV}. Springer, 2024.

\bibitem[Zou et~al.(2023)Zou, Yang, Zhang, Li, Li, Wang, Wang, Gao, and Lee]{zou2023segment}
Xueyan Zou, Jianwei Yang, Hao Zhang, Feng Li, Linjie Li, Jianfeng Wang, Lijuan Wang, Jianfeng Gao, and Yong~Jae Lee.
\newblock Segment everything everywhere all at once.
\newblock \emph{NeurIPS}, 36, 2023.

\end{thebibliography}
}


\end{document}